%% file: 00_main.tex
\pdfoutput=1
\documentclass[11pt]{article}

\usepackage[]{acl}

\usepackage{times}
\usepackage{latexsym}

\usepackage[T1]{fontenc}
\usepackage[utf8]{inputenc}

\usepackage{microtype}

\usepackage[inline]{trackchanges}
\addeditor{YJ}

\usepackage{inconsolata}
\usepackage{amsmath}
\usepackage{amssymb}
\usepackage[export]{adjustbox}
\usepackage{subcaption}
\usepackage{booktabs}
\usepackage{stfloats}
\usepackage{multirow}
\usepackage{tabularx}
\usepackage{longtable}
\usepackage{hyperref}
\usepackage{array}
\usepackage{caption}
\usepackage{graphicx}
\usepackage{xcolor}
\usepackage{pgfplotstable}
\pgfplotsset{compat=1.18}
\pgfplotstableset{
    highlight bold/.style = {
        postproc cell content/.style={
            @cell content/.add={\boldmath}{},
        },
    },
}
\usepackage{xurl}

\newcommand\shortsection[1]{\vspace{6pt}{\noindent\bf #1.}}

\newcommand\shortsectionnp[1]{\vspace{6pt}{\noindent\bf #1}}

\newcommand\doop{\mathit{do}}
\usepackage{color}

\newcommand{\occupation}[1]{{\small\textsf{#1}}}

\title{Addressing Both Statistical and Causal Gender Fairness in NLP Models}

\author{Hannah Chen, 
    Yangfeng Ji, David Evans \\
    Department of Computer Science\\
    University of Virginia\\
    Charlottesville, VA 22904\\
  \texttt{\{yc4dx,yangfeng,evans\}@virginia.edu} \\
 }

\begin{document}
\maketitle

\input{0_abstract}
\input{1_introduction}
\input{2_background}
\input{3_disparate_metrics}
\input{4_cross_evaluation}
\input{5_method}
\input{7_conclusion}
\input{8_limitations}

% \section*{Acknowledgements}

\bibliography{anthology,custom}

\input{9_appendix}

\end{document}

%% file: 0_abstract.tex
\begin{abstract}
Statistical fairness stipulates equivalent outcomes for every protected group, whereas causal fairness prescribes that a model makes the same prediction for an individual regardless of their protected characteristics. Counterfactual data augmentation (CDA) is effective for reducing bias in NLP models, yet models trained with CDA are often evaluated only on metrics that are closely tied to the causal fairness notion; similarly, sampling-based methods designed to promote statistical fairness are rarely evaluated for causal fairness. In this work, we evaluate both statistical and causal debiasing methods for gender bias in NLP models, and find that while such methods are effective at reducing bias as measured by the targeted metric, they do not necessarily improve results on other bias metrics. We demonstrate that combinations of statistical and causal debiasing techniques are able to reduce bias measured through both types of metrics.\footnote{Code for reproducing the experiments is available at: \url{https://github.com/hannahxchen/composed-debiasing}}
\end{abstract}

%% file: 1_introduction.tex
\section{Introduction}
Auditing NLP models is crucial to measure potential biases that can lead to unfair or discriminatory outcomes when models are deployed. Several methods have been proposed to quantify social biases in NLP models including intrinsic metrics that probe bias in the internal representations of the model~\citep{caliskan2017semantics,may-etal-2019-measuring,guo2021detecting} and extrinsic metrics that measure model behavioral differences across protected groups (e.g., gender and race). In this paper, we focus on extrinsic metrics as they align directly with how models are used in downstream tasks~\citep{goldfarb-tarrant-etal-2021-intrinsic,orgad-belinkov-2022-choose}.

Proposed extrinsic bias metrics can be categorized based on whether they correspond to a statistical or causal notion of fairness. A bias metric quantifies model bias based on a fairness criterion. Two common kinds of fairness criteria are statistical and causal fairness. \textit{Statistical fairness} calls for statistically equivalent outcomes for all protected groups. Statistical bias metrics estimate the difference in prediction outcomes between protected groups based on observational data~\citep{barocas-hardt-narayanan, hardt2016equality}. \textit{Causal fairness} shifts the focus from statistical association to identifying root causes of unfairness through causal reasoning~\citep{loftus2018causal}. Causal bias metrics measure the effect of the protected attribute on the model's predictions via interventions that change the value of the protected attribute. A model satisfies \emph{counterfactual fairness}, as defined by \citet{kusner2017counterfactual}, if the same prediction is made for an individual in both the actual world and in the counterfactual world in which the protected attribute is changed.

While there is no consensus on which metric is the right one to use~\citep{czarnowska2021quantifying}, most work on bias mitigation only uses a single type of metric in their evaluation. This is typically a metric that is closely connected to the proposed debiasing method. For example, counterfactual data augmentation (CDA)~\citep{lu2019gender}, has been shown to reduce bias in NLP models. However, prior works that adopt this method often evaluate only on causal bias metrics and do not include any tests using statistical bias metrics~\citep{park-etal-2018-reducing,lu2019gender,zayed2022deep,lohia2022counterfactual,wadhwa2022fairness}. We find only one exception---\citet{garg2019counterfactual} found causal debiasing exhibits some tradeoffs between statistical and causal metrics (\autoref{sec:related-work}). This raises concerns about the effectiveness and reliability of these debiasing methods in settings where multiple fairness criteria may be desired.

In this work, we first show that methods designed to reduce bias according to one fairness criteria often do not reduce bias as measured by other bias metrics. Then, we propose training methods to achieve statistical and causal fairness for gender in NLP models. We focus on gender bias as it is a well-studied problem in the literature.

\begin{figure*}[tb]
 \centering
 \includegraphics[width=\textwidth]{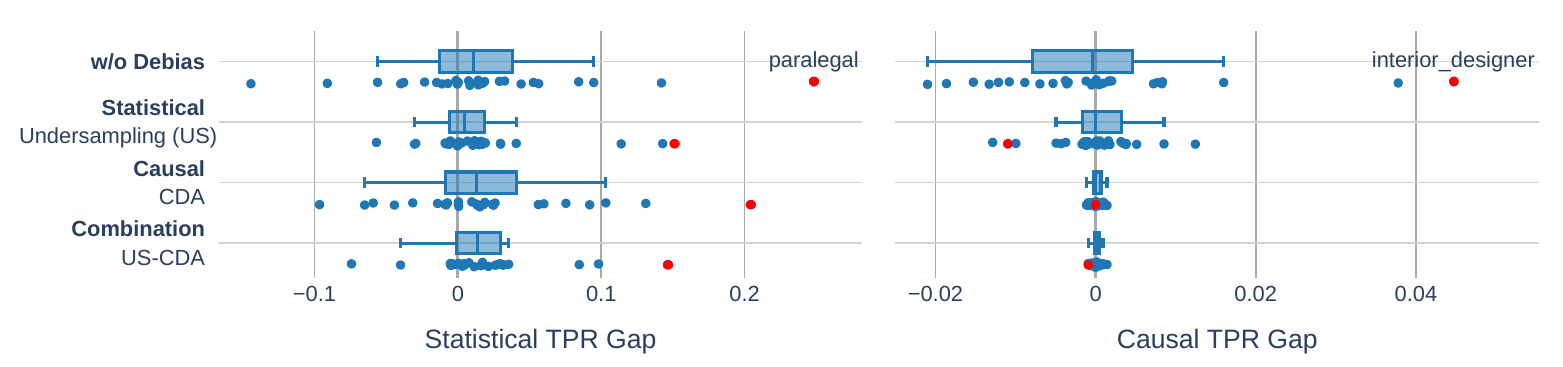}
 \caption{Statistical and causal debiasing methods perform best on the bias metric aligned with their targeted fairness notion. However, CDA is not effective at reducing statistical TPR gap. Our proposed combination approach achieves the best overall results. Results are based on BiasBios dataset with BERT-Base-Uncased model. \autoref{sec:cross-evaluation} provides details on the experiments.}
 \label{fig:cover}
\end{figure*}

\shortsection{Contributions} We empirically show the differences between statistical and causal bias metrics and explain why optimizing one of them may not improve the other (\autoref{sec:disparate-metrics}). We find that they may even disagree on which gender the model is biased towards. We cross-evaluate statistical and causal-based debiasing methods on both types of bias metrics (\autoref{sec:cross-evaluation}), and find that debiasing methods targeted to one type of fairness may even make other bias metrics worse (\autoref{sec:cross-evaluation-results}). We propose debiasing methods that combine statistical and causal debiasing techniques (\autoref{sec:method}). Our results, summarized in \autoref{fig:cover}, show that a combined debiasing method achieves the best overall results when both statistical and causal bias metrics are considered.

%% file: 2_background.tex
\section{Background}
This section provides background on bias metrics based on statistical and causal notions of fairness and overviews bias mitigation techniques.

\subsection{Bias Metrics}
\label{sec:fairness-metrics}
We consider a model fine-tuned for a classification task where the model $f$ makes predictions $\hat{Y}$ given inputs $X$ and the ground truths are $Y$.

\shortsection{Statistical bias metrics}
Statistical bias metrics quantify bias based on \emph{statistical fairness} (also known as \emph{group fairness}), which compares prediction outcomes between groups. Common statistical fairness definitions include demographic parity (DP), which requires equal positive prediction rates (PPR) for every group~\citep{barocas-hardt-narayanan}. Different from DP, equalized odds consider ground truths and demand equal true positive rates (TPR) and false positive rates (FPR) across groups~\citep{hardt2016equality}.

Statistical PPR gap ($\mathcal{SG}^{\mathsf{PPR}}$) between binary genders $g$ (female) and $\neg g$ (male) can be defined as~\citep{zayed2022deep}:
\begin{align*}
     \mathbb{E}[\hat{Y}=1 \; | \; G=g]-\mathbb{E}[\hat{Y}=1 \; | \; G=\neg g]
\end{align*}
where the model predictions $\hat{Y}$ can be either 0 or 1. If $\mathcal{SG}^{\mathsf{PPR}}>0$, the model produces positive predictions for females more often than for males.

Statistical TPR gap of binary genders for class $y$ can be formulated as~\citep{de-arteaga2019bias}:
\begin{align*}
    \mathcal{SG}^{\mathsf{TPR}}_{y} &= \mathsf{TPR}_{s}(g,y)-\mathsf{TPR}_{s}(\neg g,y) \\
    \mathsf{TPR}_{s}(g,y) &=\mathbb{E}[\hat{Y}=y \; | \; G=g,Y=y]
\end{align*}

A positive $\mathcal{SG}^{\mathsf{TPR}}$ would mean that the model outputs the correct positive prediction for female inputs more often than for male inputs. Statistical FPR gap can be defined analogously as in \autoref{eq:statistical_fpr_gap} (Appendix~\ref{app:fpr-gap-definition}).

\shortsection{Causal bias metrics}
Causality-based bias metrics for NLP models are usually based on counterfactual fairness~\citep{kusner2017counterfactual}, which requires the model to make the same prediction for the text input even when group identity terms in the input are changed. The evaluation set is usually constructed by perturbing the identity tokens in the inputs from datasets~\citep{prabhakaran-etal-2019-perturbation,garg2019counterfactual,qian-etal-2022-perturbation} or by creating synthetic sentences from templates~\citep{dixon2018measuring,lu2019gender,huang-etal-2020-reducing}.

Following \citet{garg2019counterfactual}, we can define causal gender gap for an input $x$ as:
\begin{align*}
    |f(x\,|\,\doop(G=g)) - f(x\,|\,\doop(G=\neg g))|
\end{align*}
where the $\doop$-operator enforces an intervention on gender. The term $f(x\,|\,\doop(G=g))$ indicates the model's prediction for $x$ if the gender of $x$ were set to female. To identify the bias direction, we will consider the causal gap without the absolute value. More information on how we perform gender intervention on texts is given in Appendix~\ref{app:gender-intervention}. 

Causal PPR Gap ($\mathcal{CG}^{\mathsf{PPR}}$) can be estimated by the average causal effect of the protected characteristic on the model's prediction being positive.~\citep{rubin1972estimating,Pearl2016CausalII}:
\begin{align*}
\mathbb{E}[\hat{Y}=1\,|\,\doop(G=g)]-\mathbb{E}[\hat{Y}=1\,|\,\doop(G=\neg g)]
\end{align*}
If $\mathcal{CG}^{\mathsf{PPR}}$ is zero, it would mean that gender has no influence on model's positive prediction outcome. 
To compare with statistical TPR gap, we formulate causal TPR gap by averaging the TPR difference for each individual:
\begin{align*}
    \mathcal{CG}^{\mathsf{TPR}}_{y} &= \mathsf{TPR}_{c}(g,y)-\mathsf{TPR}_{c}(\neg g,y) \\
    \mathsf{TPR}_{c}(g,y) &= \mathbb{E}[\hat{Y}=y\,|\,do(G=g), Y=y]\nonumber
\end{align*}
Similarly, we can define causal FPR gap as in \autoref{eq:causal_fpr_gap} (Appendix~\ref{app:fpr-gap-definition}).

\shortsection{Comparing statistical and causal bias metrics}
The key difference between statistical and causal metrics is how the test examples are selected and generated for evaluation. Statistical metrics are based on the original unperturbed examples, while causal metrics consider an additional perturbation process to generate test examples besides the original examples. Proponents of causal metrics argue that statistical metrics are based on observational data, which may contain spurious correlations and therefore cannot determine whether the protected attribute is the reason for the observed statistical differences~\citep{kilbertus2017avoiding,Nabi_Shpitser_2018}. On the other hand, statistical metrics are easy to assess, whereas causal metrics require a counterfactual version of each instance. Due to the discrete nature of texts, we can conveniently generate counterfactuals at the intervention level by perturbing the identity terms in the sentences~\citep{garg2019counterfactual}. Yet, it is possible to produce ungrammatical or nonsensical sentences using such perturbations~\citep{morris-etal-2020-reevaluating}. In addition, changing the identity terms alone may not be enough to hide the identity signals as there could be other terms or linguistic tendencies that are correlated with the target identity. \citet{czarnowska2021quantifying} provides a comprehensive comparison of existing extrinsic bias metrics in NLP.

\subsection{Bias Mitigation}
Bias mitigation techniques for NLP models can be categorized broadly based on whether the mitigation is done to the training data (pre-processing methods), to the learning process (in-processing), or to the model outputs (post-processing).

\shortsectionnp{Pre-processing methods} attempt to mitigate bias by modifying the training data before training. Statistical methods adjust the distribution of the training data through \emph{resampling} or \emph{reweighting}. Resampling can be done by either adding examples for underrepresented groups~\citep{dixon2018measuring,costa-jussa-de-jorge-2020-fine} or removing examples for overrepresented groups~\citep{wang2019iccv,han-etal-2022-balancing}. Reweighting assigns a weight to each training example according to the frequency of its class label and protected attribute~\citep{calders2009building,kamiran2012data,han-etal-2022-balancing}. Causal methods such as counterfactual data augmentation (CDA) augment the training set with examples substituted with different identity terms~\citep{lu2019gender}. This is the same as data augmentation based on gender swapping~\citep{zhao-etal-2018-gender,park-etal-2018-reducing}. While both statistical and causal methods seek to balance the group distribution, CDA performs interventions on the protected attribute whereas resampling and reweighing do not modify the attribute in the examples. Previous works have also considered removing protected attributes~\cite{de-arteaga2019bias}. However, this ``fairness through blindness'' approach is ineffective as there may be other proxies correlate with the protected attributes~\citep{chen2019fairness}.

\shortsectionnp{In-processing methods} incorporate a fairness constraint in the training process. The constraint can be either based on statistical fairness~\citep{kamishima2012fairness,zafar2017fairness,donini2018empirical,subramanian-etal-2021-evaluating,shen-etal-2022-optimising} or causal fairness~\citep{garg2019counterfactual}. Adversarial debiasing methods train the model jointly with a discriminator network from a typical GAN as an adversary to remove features corresponding to the protected attribute from the intermediate representations~\citep{zhang2018mitigating,elazar-goldberg-2018-adversarial,li-etal-2018-towards,han-etal-2021-diverse} 

\shortsectionnp{Post-processing methods} adjust the outputs of the model at test time to achieve desired outcomes for different groups~\citep{kamiran2010discrimination,hardt2016equality,pmlr-v65-woodworth17a}. \citet{zhao-etal-2017-men} use a corpus-level constraint during inference. \citet{ravfogel-etal-2020-null} remove protected attribute information from the learned representations.

\input{6_related}

%% file: 6_related.tex
\subsection{Related Work}
\label{sec:related-work}

\citet{garg2019counterfactual} is the only work that evaluates NLP models with both statistical and causal bias metrics. They evaluate toxicity classifiers trained with CDA and counterfactual logit pairing and observe a tradeoff between counterfactual token fairness and TPR gaps. \citet{han2023dualfair} is the only work that attempts to achieve both statistical and causal fairness through fair representational learning on tabular data. 

Previous work has studied the impossibility theorem of statistical fairness, which states that, for binary classification, equalizing multiple common statistical bias metrics between protected attributes is impossible unless the distribution of outcome is equal for both groups~\citep{kleinberg2016inherent,chouldechova2017fair,bell2023possibility}. While these works focus on tabular data and statistical bias metrics, our work studies statistical and causal bias metrics used for NLP tasks.

Comparison between various bias metrics for NLP models has also been explored. Intrinsic and extrinsic bias metrics have been shown to have no correlation with each other~\citep{delobelle-etal-2022-measuring,cabello2023independence}. \citet{delobelle-etal-2022-measuring} also shows that the measure of intrinsic bias varies depending on the choice of words and templates used for evaluation. \citet{shen-etal-2022-representational} find no correlation between statistical bias metrics and an adversarial-based bias metric, which measures the leakage of protected attributes from the intermediate representation of a model.

\citet{dwork2012fairness} proposes individual fairness, which demands similar outcomes to similar individuals. This is similar to counterfactual fairness in the sense that two similar individuals can be considered as counterfactuals of each other~\citep{loftus2018causal,pmlr-v106-pfohl19a}. The difference is that individual fairness considers similar individuals based on some distance metrics while counterfactual fairness considers a counterfactual example for each individual from a causal perspective. \citet{pmlr-v28-zemel13} proposes learning representations with group information sanitized and individual information preserved to achieve both individual and group (statistical) fairness.

%% file: 3_disparate_metrics.tex
\section{Bias Metrics Are Disparate}\label{sec:disparate-metrics}

Disparities between different statistical fairness definitions and group and individual fairness have been studied in the tabular data settings (\autoref{sec:related-work}). We focus on the most common type of bias metrics, statistical and causal, used for evaluating NLP tasks. We first explain why statistical and causal bias metrics may produce inconsistent results. We then report on the experiments to measure disparities between the metrics on evaluating gender bias in an occupation classification task.

\subsection{Statistical does not Imply Causal Fairness}
\label{sec:why-disparity}

While correlation and causation can happen simultaneously, correlation does not imply causation~\citep{fisher1958cigarettes}. Correlation refers to the statistical dependence between two variables. Statistical correlation is not causation when there is a confounding variable that influences both variables~\citep{pearl_2009}, leading to spurious correlations~\citep{pearson1896spurious}. 

To equate statistical estimates with causal estimates, the \emph{exchangeability assumption} must be satisfied~\citep{neal2015introduction}. This means that the potential outcome of a protected group is independent of the group assignment. The model's prediction outcome should be the same even when the groups are swapped. One common way to achieve this is through randomized control trials by randomly assigning individuals to different groups~\citep{fisher:1935}, making the groups more comparable. In the case of bias evaluation, it is impossible to assign gender or identity to a person randomly. Furthermore, most data are sampled from the Internet, which does not guarantee diversity and may still encode bias~\citep{bender2021stochastic}. Despite the disparities between statistical and causal bias estimation, it does not entail that achieving both statistical and causal fairness is impossible.

\subsection{Evaluation}
\label{sec:disparite-metric-evaluation}

\shortsection{Task} We use the BiasBios dataset~\citep{de-arteaga2019bias} comprising nearly 400,000 online biographies of 28 unique occupations scraped from the CommonCrawl. The task is to predict the occupation given in the biography with the occupation title removed. Each biography includes the name and the pronouns of the subject. The gender of the subject is determined by a pre-defined list of explicit gender indicators (Appendix~\ref{app:gender-intervention}). We use the train-dev-test split of the BiasBios dataset from~\citet{ravfogel-etal-2020-null}. We perform a different data pre-processing for the biographies (see Appendix~\ref{app:biasbios-dataset-construction} for details). 

\shortsection{Setup} We fine-tune ALBERT-Large~\citep{lan2020albert} and BERT-Base-Uncased~\citep{devlin2019bert} on the BiasBios dataset with normal training. We then evaluate the models with statistical and causal TPR gap.

\begin{figure}[!htb]
    \centering
    \begin{subfigure}[b]{0.9\linewidth}
    \centering
    \includegraphics[width=\textwidth]{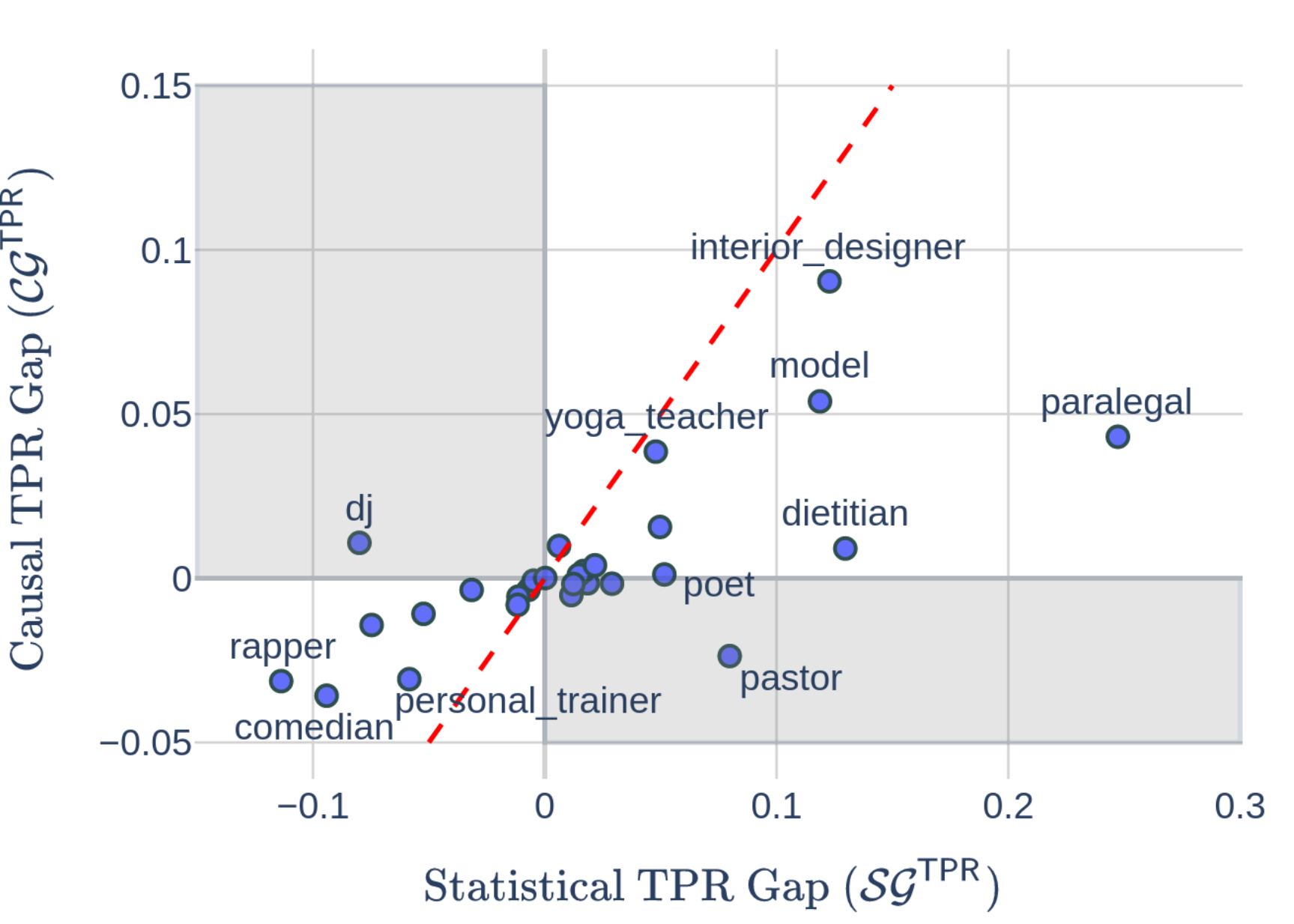}
    \caption{ALBERT-Large}
    \label{fig:disparate-metrics-albert-tpr}
    \end{subfigure}
    \vfill
    \begin{subfigure}[b]{0.9\linewidth}
    \centering
    \includegraphics[width=\textwidth]{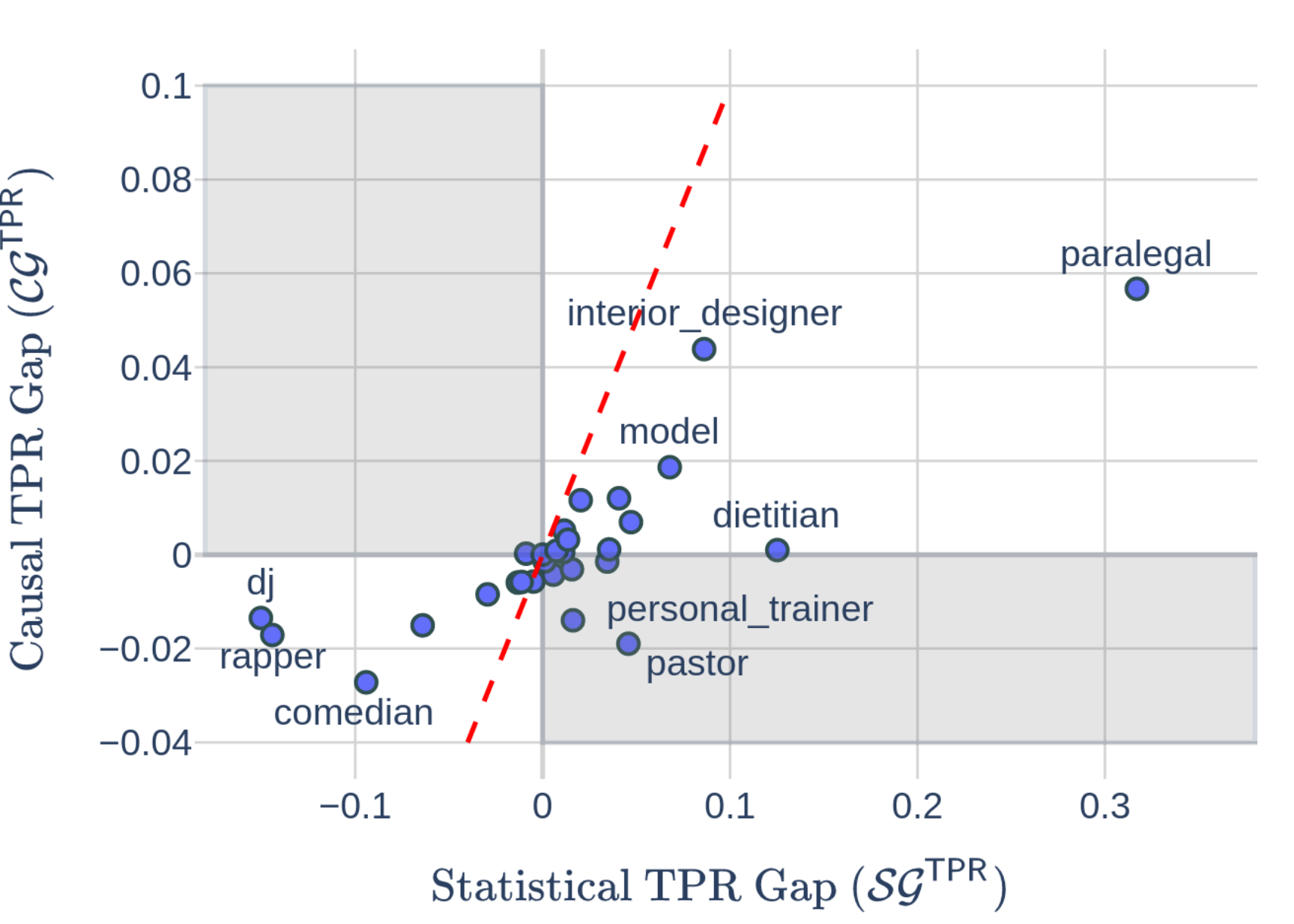}
    \caption{BERT-Base-Uncased}
    \label{fig:disparate-metrics-bert-tpr}
    \end{subfigure}
    \caption{Statistical and causal TPR gaps evaluated on models with normal training. Red dashed line indicates $\mathcal{SG}=\mathcal{CG}$. Shaded areas represent $\mathcal{SG}$ and $\mathcal{CG}$ reporting opposite gender bias direction.}
    \label{fig:disparate-metrics-tpr}
\end{figure}

\shortsection{Results} \autoref{fig:disparate-metrics-tpr} shows the statistical and causal TPR gap for ALBERT and BERT models. Each data point represents the TPR gap of an occupation evaluated over the test examples with the occupation label. The results reveal the disparity between statistical estimation and causal estimation. Most occupations are off the red dashed line where $\mathcal{SG}=\mathcal{CG}$. For nearly all occupations, $\mathcal{CG}$ is closer to zero than $\mathcal{SG}$. In addition, we find a few cases where $\mathcal{SG}$ and $\mathcal{CG}$ show bias in opposite directions such as \occupation{dj} and \occupation{pastor} in \autoref{fig:disparate-metrics-albert-tpr}. Similar results are found for statistical and causal FPR gap (see Appendix~\ref{app:disparate-metrics}).

\subsection{Bag-of-Words Analysis}
\label{sec:bow-analysis}

\begin{figure*}[!ht]
    \centering
    \includegraphics[width=\textwidth]{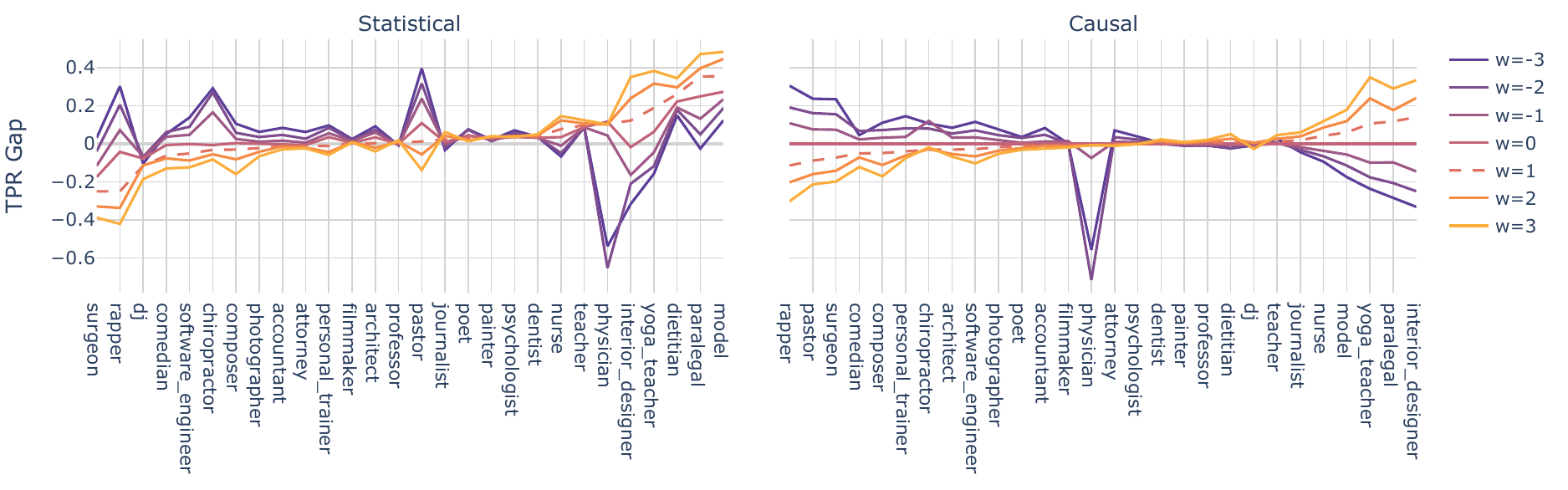}
    \caption{Statistical and causal TPR gap of BoW model per occupation when adjusting both gender token weights. $w=1$ indicates the weight is unchanged. Occupations are sorted by gap with $w=1$. Increasing the magnitude of the gender token weights increases bias on both statistical and causal bias metrics. Yet, $\mathcal{CG}^{\mathsf{TPR}}=0$ when $w=0$.}
    \label{fig:bow-gender-weight-adjusted}
\end{figure*}

To test the extent to which statistical and causal bias metrics can capture gender bias we train a Bag-of-Words (BoW) model with logistic regression on the BiasBios dataset where we can intentionally control the model's bias. We do this by identifying the model weights corresponding to gender signal tokens (Appendix~\ref{app:gender-intervention}) and multiplying the weights for these tokens by a weight $w$. This allows us to tune the bias of a simple model and see how the different bias metrics measure the resulting bias.

\autoref{fig:bow-gender-weight-adjusted} shows $\mathcal{SG}^{\mathsf{TPR}}$ and $\mathcal{CG}^{\mathsf{TPR}}$ of the BoW model when changing the weights for all gender-associated tokens. The magnitude of both bias scores increases as we increase the weighting of the gender tokens. The model is biased in the opposite gender direction when we reverse the weight $w$ by multiplying by a negative value. This demonstrates that both metrics are indeed able to capture bias in the model and, for the most part, reflect the amount of bias in the expected direction. Note that  $\mathcal{CG}^{\mathsf{TPR}}=0$ for all occupations when $w=0$. This is because $\mathcal{CG}^{\mathsf{TPR}}$ considers the average difference between pairs of sentences that only differ in tokens representing the gender. When $w=0$, the model would exclude all gender tokens and each sentence pair would render the same to the model. On the other hand, $\mathcal{SG}^{\mathsf{TPR}}$ is nonzero for most occupations when $w=0$, meaning that it captures gender bias beyond explicit gender indicators. This suggests models trained to achieve causal fairness may still be biased toward other implicit gender features not identified in our explicit gender token list.

The spikes in \autoref{fig:bow-gender-weight-adjusted} may be attributed to the relatively large gap in token weights between the two genders for predicting the occupation, as shown in \autoref{fig:bow-gender-token-weight-sum}. The increased TPR gap is particularly significant for occupations with positive token weights for the dominant gender and negative token weights for the other gender, such as \occupation{rapper} and \occupation{paralegal}. In one extreme case, both gender token weights are positive for \occupation{physician}, with female tokens having a lot higher weight value than male tokens. This results in a huge TPR gap increase only in the negative direction when applying a larger negative value of $w$.

\begin{figure*}[!ht]
    \centering
    \includegraphics[width=\linewidth]{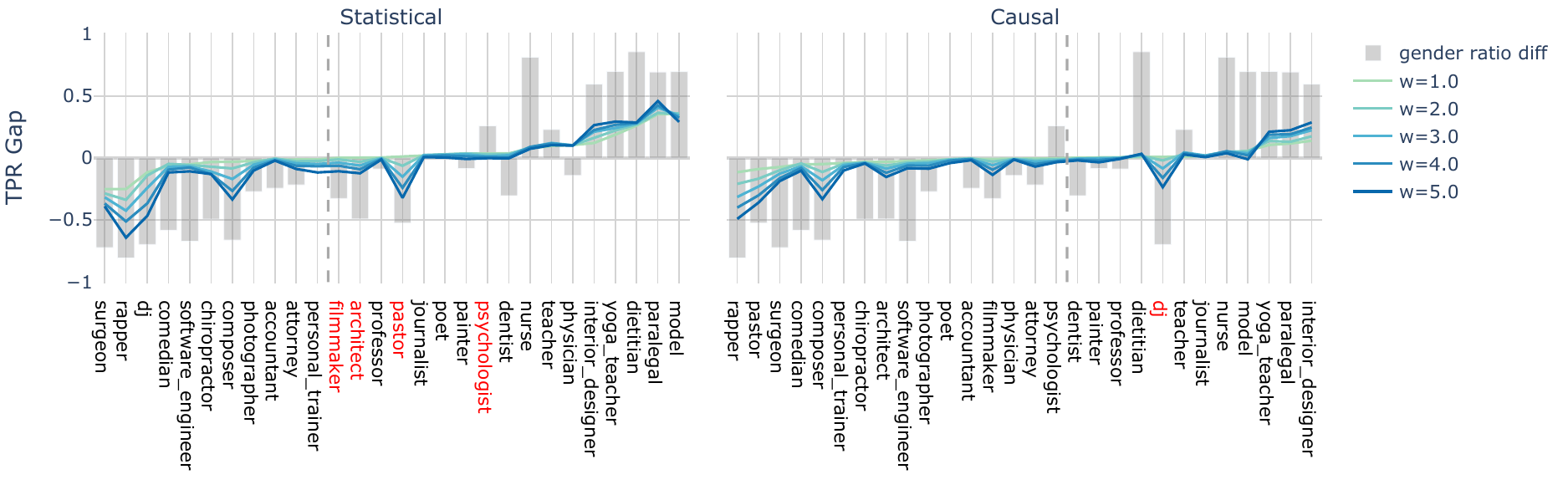}
    \caption{Statistical and causal TPR gap of BoW model per occupation when increasing female token weights in the model. The occupations highlighted in red demonstrate an increased TPR gap toward the opposite bias direction. The grey dashed line shows where the gap is zero when $w=1$. The grey bars are the gender ratio difference of the occupation in the training set.}
    \label{fig:bow-increase-female-weight}
\end{figure*}

We further analyze how model weights of individual gender affect bias scores. \autoref{fig:bow-increase-female-weight} shows the statistical and causal TPR gap of each occupation when increasing female token weights, and \autoref{fig:bow-increase-male-weight} (in  Appendix~\ref{app:bow-analysis}) shows the results of increasing male token weights. We observed that increasing female token weights has a greater effect on increasing the TPR gap of male-biased occupations (on the left side of the grey dashed line in \autoref{fig:bow-increase-female-weight}), and vice versa. In addition, some occupations (as highlighted in red) show an increased TPR gap to the opposite gender bias direction of their bias scores indicated by the metric when $w=1$. For instance, \occupation{filmmaker}, \occupation{architect}, and \occupation{pastor} are female-biased based on the statistical metric but become male-biased when increasing the female token weights due to their negative weight values (\autoref{fig:bow-gender-token-weight-sum}). We find that these occupations are the ones that the two metrics contradict in the bias direction (\autoref{tab:bow-bias-contradiction}). However, both metrics show similar patterns and directions of TPR gap increase across occupations (\autoref{fig:bow-tpr-gap-change}). The only difference is the starting point of TPR gap score when $w=1$.

%% file: 4_cross_evaluation.tex
\section{Cross-Evaluation}\label{sec:cross-evaluation}
This section cross-evaluates the effectiveness of existing debiasing methods on gender bias in an occupation classification and toxicity detection task. We show using statistical and causal debiasing methods alone may not achieve both types of fairness.

\subsection{Setup}
We focus on pre-processing methods since \citet{shen-etal-2022-optimising} found that resampling and reweighting achieve better statistical fairness than the in-processing and post-processing methods. For the statistical methods, we apply both resampling using oversampling (OS) and undersampling (US) and reweighting (RW) using the weight calculation from~\citet{kamiran2012data}. For the causal methods, we fine-tune the model with CDA. %We also include results with normal training without debiasing. 

We apply each debiasing method to the ALBERT-Large~\citep{lan2020albert} and BERT-Base-Uncased~\citep{devlin2019bert} models. We also include experiments with Zari~\citep{webster2020measuring}, which is an ALBERT-Large model pre-trained with CDA. To consider the effect of CDA during pre-training alone and during both pre-training and fine-tuning, we fine-tune Zari with normal training and CDA. Training details are provided in Appendix~\ref{app:train-details}.

\begin{figure*}[tb]
 \centering
 \includegraphics[width=\textwidth]{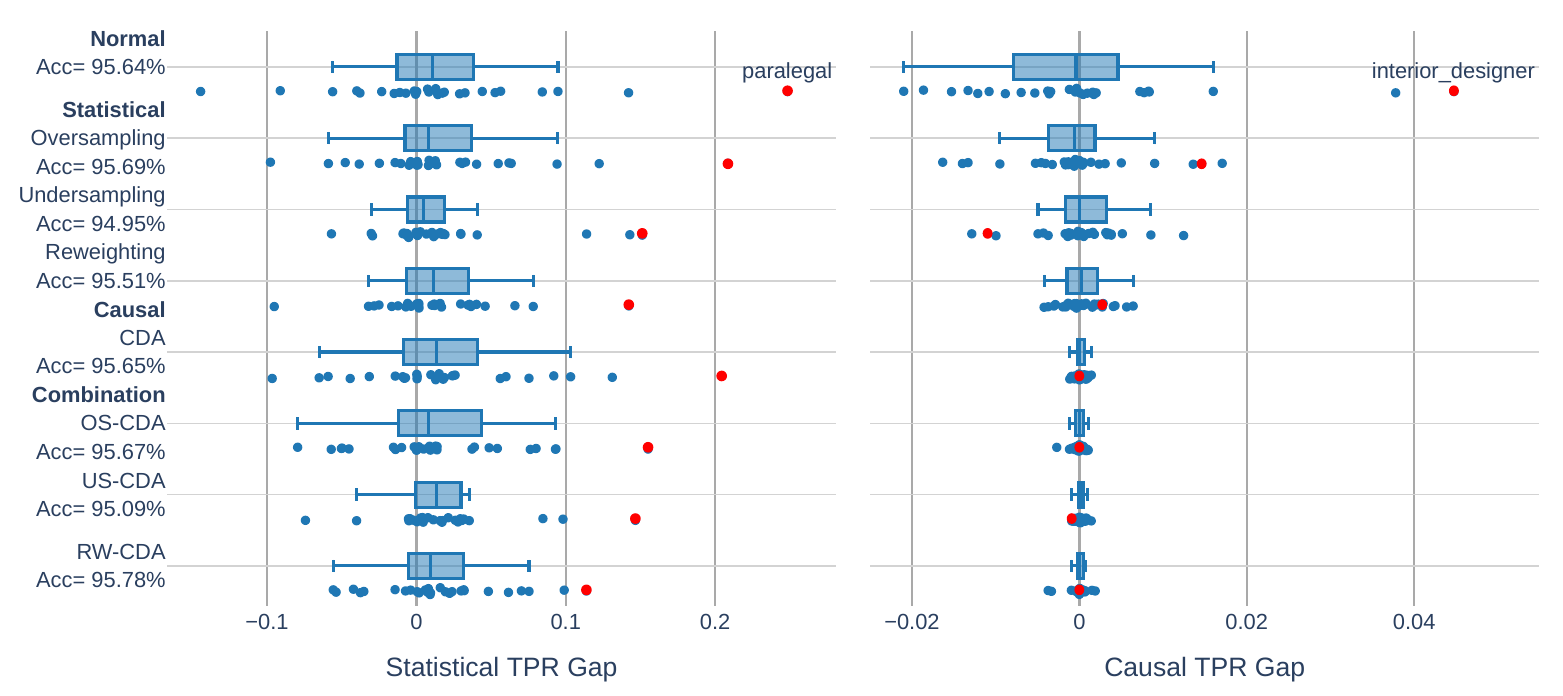}
 \caption{Statistical and causal TPR gap per occupation evaluated on BERT-Base-Uncased model, averaged over 3 different runs. Each data point is computed over test examples labeled with the same occupation. We show outliers for normal training in red dots and how their values change with different debiasing methods. Statistical and causal debiasing methods perform better on the metric they are targeting, but may not reduce bias on the other metric. Our proposed methods, \texttt{US-CDA} and \texttt{RW-CDA}, achieve the best overall performance.}
 \label{fig:bert-sg-vs-cg-tpr}
\end{figure*}

\subsection{Tasks}

We test all the models on two benchmark tasks for bias detection: occupation classification and toxicity detection.

\shortsection{Occupation Classification} We use the BiasBios dataset introduced in \autoref{sec:disparite-metric-evaluation}. We evaluate gender bias with TPR and FPR gap based on both statistical and causal notions of fairness as defined in \autoref{sec:fairness-metrics}. Since the BiasBios dataset contains multiple classes, we follow \citet{romanov-etal-2019-whats} and compute a single score that quantifies overall gender bias. For each bias metric $\mathit{M}$ (e.g., $\mathcal{SG}^{\mathsf{TPR}}_{g,y}$), we compute the root mean square of the bias score across all occupation classes $Y$:
\begin{align*}
    \mathit{RMS}_{\mathit{M}}=\sqrt{\frac{1}{|Y|}\sum_{y\in Y} (\mathit{M}_{y})}
\end{align*}
where $\mathit{M}_{y}$ is the bias score for occupation $y$ computed with $\mathit{M}$.

\shortsection{Toxicity Detection} We use the Jigsaw dataset consisting of approximately 1.8M comments taken from the Civil Comments platform. The task is to predict the toxicity score of each comment. For our experiments, we use binary toxicity labels, toxic and non-toxic.  In addition to the toxicity score, a subset of examples are labeled with the identities mentioned in the comment. We only select the examples labeled with female and male identities and with high annotator agreement on the gender identity labels. Since some examples contain a mix of genders, we assign the gender to each example based on the gender labeled with the highest agreement. To perform gender intervention with CDA, we use the \textit{gender-bender} Python package to generate counterfactual examples~\footnote{\url{{https://github.com/Garrett-R/gender_bender}}}. Appendix~\ref{app:jigsaw-dataset-construction} provides details on how we preprocess the data. Following \citet{zayed2022deep}, we compute statistical and causal PPR gap. As female and male groups do not have the same label distribution, the PPR gap of a perfect predictor will be non-zero. Therefore, we also compute statistical and causal TPR gap for toxic and non-toxic classes.

\subsection{Results}
\label{sec:cross-evaluation-results}

\begin{figure*}[tb]
 \centering
 \includegraphics[width=\linewidth]{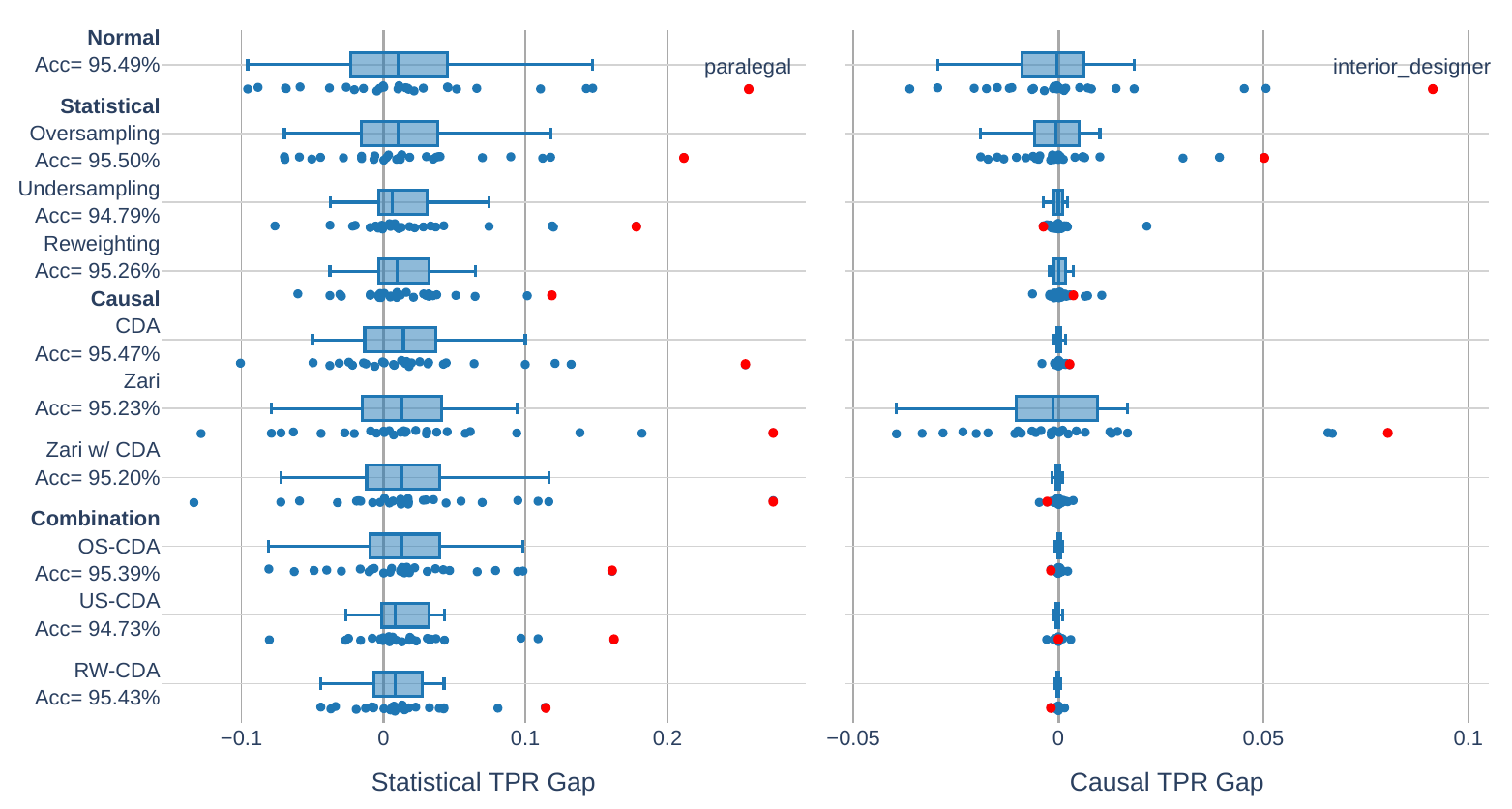}
 \caption{Statistical and causal TPR gap per occupation results for ALBERT-Large, averaged over 3 different runs.}
\label{fig:albert-sg-vs-cg-tpr}
\end{figure*}

\shortsection{Occupation classification} \autoref{fig:bert-sg-vs-cg-tpr} and \autoref{fig:albert-sg-vs-cg-tpr} show statistical and causal TPR gap per occupation evaluated on BERT and ALBERT models with each debiasing method. Causal debiasing methods show greater effectiveness when evaluated with the causal metric (we discuss the combination methods included in these figures in \autoref{sec:method}). Fine-tuning with CDA reduces $ \mathcal{CG}^{\mathsf{TPR}}$ to nearly zero for all occupations, but does not produce any significant reduction for $ \mathcal{SG}^{\mathsf{TPR}}$. On the other hand, Zari exhibits higher statistical and causal gap than performing CDA during fine-tuning (\autoref{fig:albert-sg-vs-cg-tpr}). Thus, using CDA during pre-training alone is insufficient to reduce bias. Statistical debiasing methods such as undersampling and reweighting reduce bias on both statistical and causal metrics, though the bias reduction on the causal metric is not as significant as CDA. We find that oversampling is less effective than other statistical debiasing methods on both metrics. We found similar results with statistical and causal FPR gaps (Appendix~\ref{app:sg-vs-cg-fpr}).

\shortsection{Toxicity detection}
\autoref{tab:jigsaw-bert} shows the bias evaluation results for the BERT model trained with different debiasing methods on the Jigsaw dataset. We find that statistical and causal bias metrics sometimes disagree on which gender the model is biased toward. Similar to the results for the BiasBios task, statistical and causal debiasing methods do particularly well on the bias metrics based on their targeted fairness definition. However, they increase bias on metrics that use the other type of fairness notion. Similar results are found for ALBERT model (Appendix~\ref{app:jigsaw-bias-score-albert}).

\input{tables/jigsaw-ppr-gap-bert}

%% file: tables/jigsaw-ppr-gap-bert.tex
\begin{table*}[!ht]
    \centering
    \resizebox{\linewidth}{!}{%
    \begin{filecontents*}{csv_files/jigsaw-ppr-gap-bert.csv}
method,auc,auc_std,statistical_ppr_gap,statistical_ppr_gap_std,causal_ppr_gap,causal_ppr_gap_std,statistical_tpr_gap,statistical_tpr_gap_std,causal_tpr_gap,causal_tpr_gap_std,statistical_tpr_neg_gap,statistical_tpr_neg_gap_std,causal_tpr_neg_gap,causal_tpr_neg_gap_std
Normal,0.9250367107195302,0.003485778992795655,-2.7902403851777047,0.27786430471208146,0.8920704845814975,0.10357368560694431,-2.773905410987194,0.6676681073417581,2.333931777378815,1.0590938027829533,1.2753542117400563,0.3003562022505903,-0.7278675117562871,0.11017946093075022
CDA,0.9279735682819383,0.0024914654186427606,-3.017147370514064,0.23490275162784718,0.25330396475770955,0.07647821826725237,-2.6244320528707155,2.0715399221886597,0.3590664272890485,0.5677338707663159,1.5161935160125317,0.29104081315941993,-0.24125945614393782,0.05981900976621465
OS,0.9317180616740088,0.0016293936138892851,-1.209145906420477,0.22252071783986643,1.3252569750367107,0.3102870280340005,2.2100888062783963,0.3545601476490087,5.242369838420108,0.41564943995656106,0.19631776090467623,0.17284167034439785,-0.8791658147618074,0.30408493056762553
US,0.9244860499265786,0.003942943195917029,-1.5440463060877536,0.26499691802974645,1.6666666666666659,0.28704871956423794,1.6101301115241642,1.1101843196136325,4.560143626570916,0.6281097193729226,0.37023593093179485,0.23573362551367275,-1.337149867102842,0.2562167278318633
RW,0.9290014684287812,0.002683651524595759,-1.4394929612737617,0.308091046775431,1.4427312775330399,0.2434006527790133,2.093143329202811,0.849970779250391,4.919210053859964,0.5277177183733955,0.3853895026639531,0.2584537496599199,-1.0468206910652218,0.25087485831409473
OS-CDA,0.9310205580029368,0.0016597143253241394,-2.089619315027118,0.3030007486929356,0.17621145374449365,0.1598497874968964,-1.11188558446923,0.991578511244936,0.3949730700179533,0.4598293886845853,0.7906215809346717,0.28168187840919934,-0.15129830300552033,0.14720915968104684
US-CDA,0.9305433186490456,0.0022468639865050343,-1.9004842929865922,0.18717726853174252,0.11380323054331898,0.11338799714743356,-1.6591800908715415,1.878341681468191,0.1436265709156194,0.7036236244978609,0.5735871976483709,0.2648222012745839,-0.11040686976078509,0.06147330353045556
RW-CDA,0.9303964757709251,0.0026457199334401337,-1.755500499154087,0.3638498699330784,0.3267254038179149,0.11037662539187147,0.5589116067740618,1.2735117476906126,1.0771992818671454,0.7445759911428266,0.6192878472536423,0.4047435994928218,-0.2412594561439379,0.10049646884027404
\end{filecontents*}
\pgfplotstabletypeset[
        col sep = comma,
        columns={method, sg-ppr, cg-ppr, sg-tpr, cg-tpr, sg-tpr-2, cg-tpr-2},
        columns/method/.style={column name=Method, string type, column type/.add={}{|}},
        columns/sg-ppr/.style={string type, column name=$\mathcal{SG}^{\mathsf{PPR}}$},
        columns/cg-ppr/.style={string type, column name=$\mathcal{CG}^{\mathsf{PPR}}$},
        columns/sg-tpr/.style={string type, column name=$\mathcal{SG}^{\mathsf{TPR}}_{y=1}$},
        columns/cg-tpr/.style={string type, column name=$\mathcal{CG}^{\mathsf{TPR}}_{y=1}$},
        columns/sg-tpr-2/.style={string type, column name=$\mathcal{SG}^{\mathsf{TPR}}_{y=0}$},
        columns/cg-tpr-2/.style={string type, column name=$\mathcal{CG}^{\mathsf{TPR}}_{y=0}$},
        create on use/auc-mixed/.style={
            create col/assign/.code={%
              \edef\entry{\noexpand\pgfmathprintnumber[fixed, fixed zerofill, precision=3]{\thisrow{auc}}$\pm$\noexpand\pgfmathprintnumber[fixed, fixed zerofill, precision=3]{\thisrow{auc_std}}}%
              \pgfkeyslet{/pgfplots/table/create col/next content}\entry
            }
          },
        create on use/sg-ppr/.style={
            create col/assign/.code={%
              \edef\entry{\noexpand\pgfmathprintnumber[fixed, fixed zerofill, precision=2]{\thisrow{statistical_ppr_gap}}$\pm$\noexpand\pgfmathprintnumber[fixed, fixed zerofill, precision=2]{\thisrow{statistical_ppr_gap_std}}}%
              \pgfkeyslet{/pgfplots/table/create col/next content}\entry
            }
          },
        create on use/cg-ppr/.style={
            create col/assign/.code={%
              \edef\entry{\noexpand\pgfmathprintnumber[fixed, fixed zerofill, precision=2]{\thisrow{causal_ppr_gap}}$\pm$\noexpand\pgfmathprintnumber[fixed, fixed zerofill, precision=2]{\thisrow{causal_ppr_gap_std}}}%
              \pgfkeyslet{/pgfplots/table/create col/next content}\entry
            }
          },
        create on use/sg-tpr/.style={
            create col/assign/.code={%
              \edef\entry{\noexpand\pgfmathprintnumber[fixed, fixed zerofill, precision=2]{\thisrow{statistical_tpr_gap}}$\pm$\noexpand\pgfmathprintnumber[fixed, fixed zerofill, precision=2]{\thisrow{statistical_tpr_gap_std}}}%
              \pgfkeyslet{/pgfplots/table/create col/next content}\entry
            }
          },
        create on use/cg-tpr/.style={
            create col/assign/.code={%
              \edef\entry{\noexpand\pgfmathprintnumber[fixed, fixed zerofill, precision=2]{\thisrow{causal_tpr_gap}}$\pm$\noexpand\pgfmathprintnumber[fixed, fixed zerofill, precision=2]{\thisrow{causal_tpr_gap_std}}}%
              \pgfkeyslet{/pgfplots/table/create col/next content}\entry
            }
          },
        create on use/sg-tpr-2/.style={
            create col/assign/.code={%
              \edef\entry{\noexpand\pgfmathprintnumber[fixed, fixed zerofill, precision=2]{\thisrow{statistical_tpr_neg_gap}}$\pm$\noexpand\pgfmathprintnumber[fixed, fixed zerofill, precision=2]{\thisrow{statistical_tpr_neg_gap_std}}}%
              \pgfkeyslet{/pgfplots/table/create col/next content}\entry
            }
          },
        create on use/cg-tpr-2/.style={
            create col/assign/.code={%
              \edef\entry{\noexpand\pgfmathprintnumber[fixed, fixed zerofill, precision=2]{\thisrow{causal_tpr_neg_gap}}$\pm$\noexpand\pgfmathprintnumber[fixed, fixed zerofill, precision=2]{\thisrow{causal_tpr_neg_gap_std}}}%
              \pgfkeyslet{/pgfplots/table/create col/next content}\entry
            }
          },
        every head row/.style={before row={\toprule},after row=\midrule},
        every last row/.style={after row=\bottomrule},
        every row 2 column 1/.style={highlight bold},
        every row 6 column 2/.style={highlight bold},
        every row 7 column 3/.style={highlight bold},
        every row 6 column 4/.style={highlight bold},
        every row 2 column 5/.style={highlight bold},
        every row 6 column 6/.style={highlight bold},
        every row 5 column 0/.style={postproc cell content/.style=
        {@cell content=\texttt{OS-CDA}}},
        every row 6 column 0/.style={postproc cell content/.style=
        {@cell content=\texttt{US-CDA}}},
        every row 7 column 0/.style={postproc cell content/.style=
        {@cell content=\texttt{RW-CDA}}}
    ]
    {csv_files/jigsaw-ppr-gap-bert.csv}%
    }
    \caption{Bias evaluation results evaluated on the Jigsaw dataset with BERT-Base-Uncased model. The results shown are averaged over 5 different runs. All values are on a log scale with base $10^{-2}$.}
    \label{tab:jigsaw-bert}
\end{table*}

%% file: 5_method.tex
\section{Achieving Both Statistical and Causal Fairness}\label{sec:method}
In the previous section, we saw that using either statistical or causal debiasing method alone may not achieve both statistical and causal fairness. To counter this problem, this section considers simple methods that combine both statistical and causal debiasing techniques.

\subsection{Composed Debiasing Methods}\label{sec:combo-method}
We introduce three approaches that combine techniques from both statistical and causal debiasing:

\shortsection{Resampling with CDA} \texttt{OS-CDA} and \texttt{US-CDA} combines resampling methods (oversampling and undersampling) with CDA. For Biasbios, we first perform resampling on the training set, then augment the resampled set with CDA. For Jigsaw, we balance the original examples based on the original gender and the counterfactual examples based on the counterfactual gender. 

\shortsection{Reweighting with CDA} \texttt{RW-CDA} applies CDA on the training set and fine-tunes the model with reweighting. For BiasBios, we use the same weight computed on the original training set for both the original and its counterfactual pair. For Jigsaw, we use weight of 1 for all counterfactual examples.

We use different combination strategies for the two datasets as we noticed the methods used for BiasBios do not work well on the Jigsaw dataset. This may be due to the mix of genders in a subset of examples in the Jigsaw dataset. The gender signals in the examples may be flipped after performing CDA. We provide performance comparisons between the different combination strategies we have tried on the Jigsaw task in Appendix~\ref{app:combination-strategy-comparison}.

\subsection{Results}
\autoref{fig:bert-sg-vs-cg-tpr} and \autoref{fig:albert-sg-vs-cg-tpr} show statistical and causal TPR gap per occupation evaluated on the BiasBios dataset for BERT and ALBERT models. The combined methods \texttt{US-CDA} and \texttt{RW-CDA} are more effective at reducing bias on both metrics compared to other methods. To compare overall performance, we show the root mean square of each bias metric in \autoref{tab:rms-albert} and \autoref{tab:rms-bert} (both in Appendix~\ref{app:rms}). All three combination approaches perform better on $\mathcal{CG}^{\mathsf{TPR}}$ compared to using a statistical or causal debiasing method alone. \texttt{OS-CDA} and \texttt{US-CDA} also reduce bias on $\mathcal{SG}^{\mathsf{TPR}}$ (11--16\% decrease) and $ \mathcal{SG}^{\mathsf{FPR}}$ (1--8\% decrease), comparing to their statistical debiasing counterparts. \texttt{RW-CDA} achieves comparable performance on $\mathcal{SG}$ to reweighting. Undersampling and \texttt{US-CDA} sacrifice the general performance with a decrease of around 0.7\% in accuracy compared to other methods, which preserve the baseline accuracy within 0.3\%. 

\autoref{tab:jigsaw-bert} and \autoref{tab:jigsaw-albert} (Appendix~\ref{app:jigsaw-bias-score-albert}) report the results of BERT and ALBERT models for the Jigsaw dataset. While statistical and causal debiasing methods only improve one type of bias metric and worsen the other, our proposed combination approaches are able to reduce bias on both types of bias metrics. The combined methods \texttt{OS-CDA} and \texttt{US-CDA} perform better than CDA on all causal bias metrics. \texttt{RW-CDA} performs better on $\mathcal{SG}$ but is less effective at reducing bias on $\mathcal{CG}$ compared to the other combination approaches.

%% file: 7_conclusion.tex
\section{Summary}
We demonstrate the disparities between statistical and causal bias metrics and provide insight into how and why optimizing based on one type of metric does not necessarily improve the other. We show this by cross-evaluating existing statistical and causal debiasing methods on both metrics and find that they sometimes may even worsen the other type of bias metrics. To obtain models that perform well on both types of bias metrics, we introduce simple debiasing strategies that combine both statistical and causal debiasing techniques. 
%We show that they are more effective at reducing bias on both statistical and causal metrics.

%% file: 8_limitations.tex
\section*{Limitations}
Due to the limited benchmark datasets compatible with extrinsic metrics~\citep{orgad-belinkov-2022-choose}, we only conduct experiments on two gender bias tasks. Further testing is needed to determine if the bias metric disparities are present in other tasks and whether our proposed debiasing methods can still be effective. The gender intervention method used for counterfactual data augmentation is based on a predefined list of gender tokens, which may not cover all possible tokens representing gender. In addition, our experiments exclusively focus on binary-protected attributes. Future work should explore how to generalize our results to tasks with non-binary protected attributes. While our proposed debiasing methods are able to reduce bias on both statistical and causal bias metrics, there is room for improvements in the statistical bias metrics when compared to statistical debiasing methods. Future work could consider other types of debiasing techniques beyond pre-processing-based methods. For instance, in-processing methods can be adapted by enforcing both statistical and causal fairness constraints during training.

%% file: 9_appendix.tex
\appendix
\label{sec:appendix}
\onecolumn

\section{False Positive Rate Gap}
\label{app:fpr-gap-definition}

Statistical FPR gap between binary gender $g$ (female) and $\neg g$ (male) for class $y$ is defined as:
\begin{align}
    \mathcal{SG}^{\mathsf{FPR}}_{y} &= \mathsf{FPR}_{s}(g,y)-\mathsf{FPR}_{s}(\neg g,y) \label{eq:statistical_fpr_gap} \\
    \mathsf{FPR}_{s}(g,y) &=\mathbb{E}[\hat{Y}=y \; | \; G=g,Y\not=y]\nonumber
\end{align}

\noindent Causal FPR gap is computed by averaging the FPR difference for each individual:
\begin{align}
    \mathcal{CG}^{\mathsf{FPR}}_{y} &= \mathsf{FPR}_{c}(g,y)-\mathsf{FPR}_{c}(\neg g, y)\label{eq:causal_fpr_gap} \\
    \mathsf{TPR}_{c}(g,y) &= \mathbb{E}[\hat{Y}=y\,|\,do(G=g), Y\not=y]\nonumber
\end{align}

\input{appendix/dataset_details}
\clearpage
\input{appendix/metric_disparity}
\input{appendix/train_details}
\clearpage
\input{appendix/biasbios_results}
\input{appendix/jigsaw_results}

%% file: appendix/dataset_details.tex
\section{BiasBios Dataset Details}

\subsection{Dataset Statistics}
\begin{figure}[htb]
    \centering
    \includegraphics[width=0.65\linewidth]{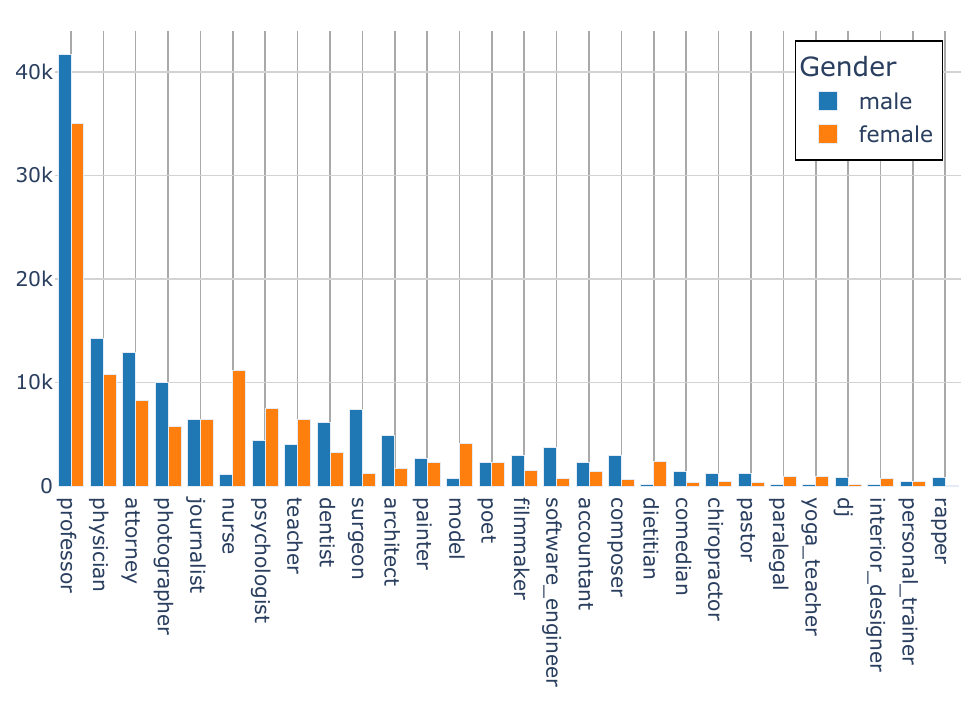}
    \caption{Gender frequency for each occupation in the training set.}
    \label{fig:biasbios-stats}
\end{figure}

The dataset contains 255,707 training examples, 39,369 validation examples, and 98,339 testing examples. Figure~\ref{fig:biasbios-stats} shows the full list of occupations and their gender frequency in the BiasBios training set. The gender and occupation distribution for validation and testing sets are similar to the training set.

\subsection{Dataset Construction}
\label{app:biasbios-dataset-construction}
The original BiasBios dataset consists of extracted biographies with the first sentences removed from each biography as they include the occupation titles corresponding to the ground truth labels. We notice a lot of the important information is in the first sentences and it is hard to correctly identify the occupation of some examples without the first sentences even for humans. Thus, we keep the first sentence but replace any occupation tokens that appear in the biography with an underscore (e.g., "Alice is a nurse working at a hospital" to "Alice is a \_ working at a hospital"). We notice that our model performance is higher than the same model trained on the original dataset~\citep{webster2020measuring}. This can be attributed to having longer sequences and more context information in the inputs.

\subsection{Gender Intervention}
\label{app:gender-intervention}
To perform gender intervention, we first identify words with explicit gender indicators in the input. If the assigned gender value is different from the original input, we swap the identified words with the corresponding words in the mapping with an opposite gender. We use the same list of explicit gender indicators used in BiasBios dataset and perform gender mapping as follows:
\begin{itemize}
    \item Bidirectional: \textcolor{blue}{he} $\leftrightarrow$ \textcolor{red}{she}, \textcolor{blue}{himself} $\leftrightarrow$ \textcolor{red}{herself}, \textcolor{blue}{mr} $\leftrightarrow$ \textcolor{red}{ms}
    \item Unidirectional: \textcolor{red}{hers} $\rightarrow$ \textcolor{blue}{his}, \textcolor{blue}{his} $\rightarrow$ \textcolor{red}{her}, \textcolor{blue}{him} $\rightarrow$ \textcolor{red}{her}, \textcolor{red}{her} $\rightarrow$ \textcolor{blue}{his} or \textcolor{blue}{him}, \textcolor{red}{mrs} $\rightarrow$ \textcolor{blue}{mr}
\end{itemize}

Words in \textcolor{blue}{blue} are associated with male gender and words in \textcolor{red}{red} are associated with female gender. Since "her" can be mapped to either "his" or "him" depending on the context, we use Part-of-Speech tagging to determine which one to map to.

\section{Jigsaw Dataset Details}

\subsection{Dataset Construction}
\label{app:jigsaw-dataset-construction}

Each comment is associated with a toxicity label and several identity labels. The label values range from 0.0 to 1.0 representing the percentage of annotators who agreed that the label fit the comment. We binarized the toxicity values and considered comments as toxic if their toxicity values exceeded 0.5. We assigned female gender to an example if its female identity label value is higher than the male one and assigned male gender vice versa. To make better differentiation between the two genders, we filtered out examples if the difference between male and female label values is smaller or equal to 0.5. We use {\fontfamily{pcr}\selectfont train.csv} from the Kaggle competition for training and validation with an 80/20 split. We use {\fontfamily{pcr}\selectfont test\_public\_expanded.csv} and {\fontfamily{pcr}\selectfont test\_private\_expanded.csv} for testing.

\input{tables/jigsaw-data-distribution}

\subsection{Dataset Statistics}
\begin{figure}[htb]
    \centering
    \includegraphics[width=\linewidth]{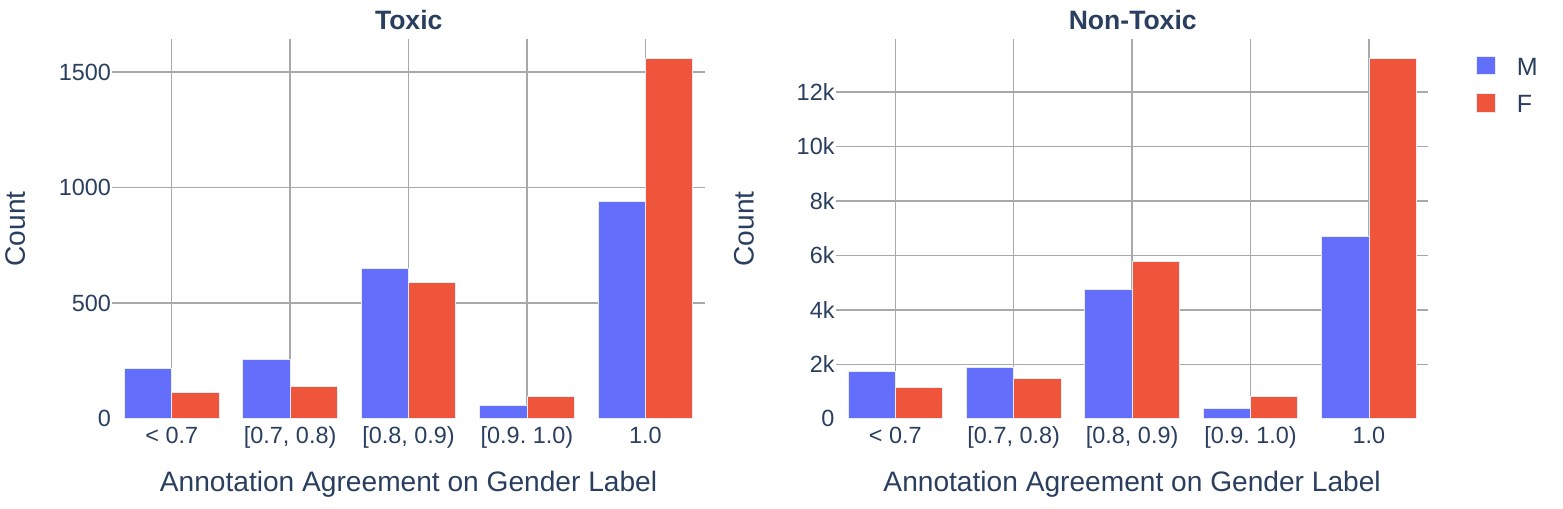}
    \caption{Distribution of annotation agreement on the gender labels. 1.0 indicates all annotators agree that the gender is mentioned in the comment.}
    \label{fig:jigsaw-gender-stats}
\end{figure}

The final dataset after pre-processing contains 42,523 training examples, 10,631 validation examples, and 5,448 testing examples. \autoref{tab:jigsaw-data-distribution} shows the gender and label distribution on the training set. All three data splits have similar distributions. We also show the distribution of the gender label values in \autoref{fig:jigsaw-gender-stats}. For examples that contain a mix of both female and male genders, we show the gender label value of the final gender we assigned (the gender with a higher label value).

%% file: tables/jigsaw-data-distribution.tex
\begin{table}[htb]
    \centering
    \begin{tabular}{c c c c}
    \toprule
    Label & Gender & Count & Percentage (\%) \\
    \midrule
    Toxic & F & 2504 & 5.89 \\
    Toxic & M & 2123 & 4.99 \\
    Non-Toxic & F & 22,465 & 52.83 \\
    Non-Toxic & M & 15,431 & 26.29 \\
    \bottomrule
    \end{tabular}
    \caption{Gender and label distribution of Jigsaw training set.}
    \label{tab:jigsaw-data-distribution}
\end{table}

%% file: appendix/metric_disparity.tex
\section{Disparities between Statistical and Causal Bias Metrics}
\label{app:disparate-metrics}

\subsection{Statistical vs Causal FPR Gap}

\begin{figure*}[!h]
    \centering
    \begin{subfigure}[b]{0.49\linewidth}
         \centering
         \includegraphics[width=\linewidth]{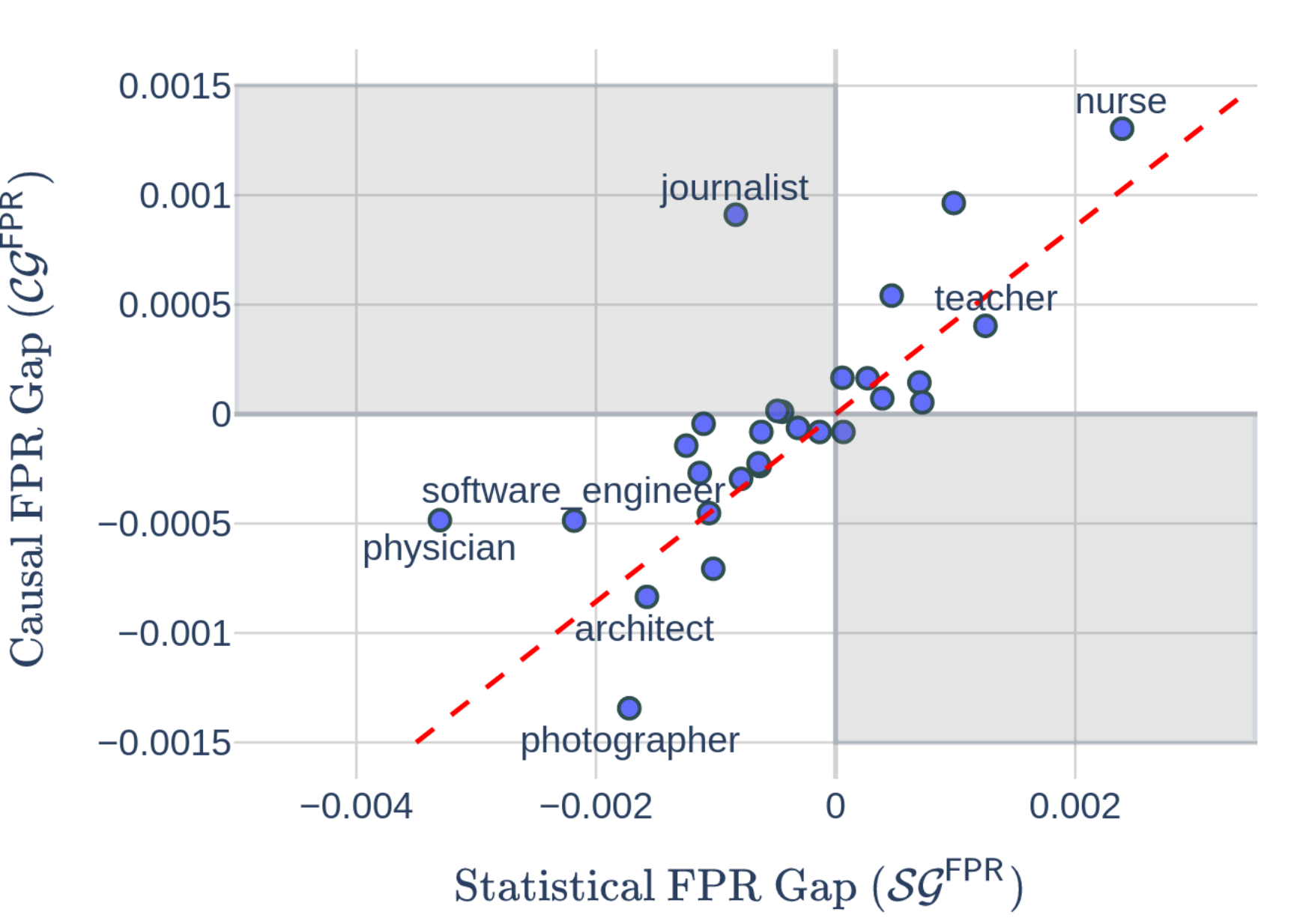}
         \caption{ALBERT-Large}
         \label{fig:albert-baseline-fpr}
     \end{subfigure}
     \hfill
     \begin{subfigure}[b]{0.49\linewidth}
         \centering
         \includegraphics[width=\textwidth]{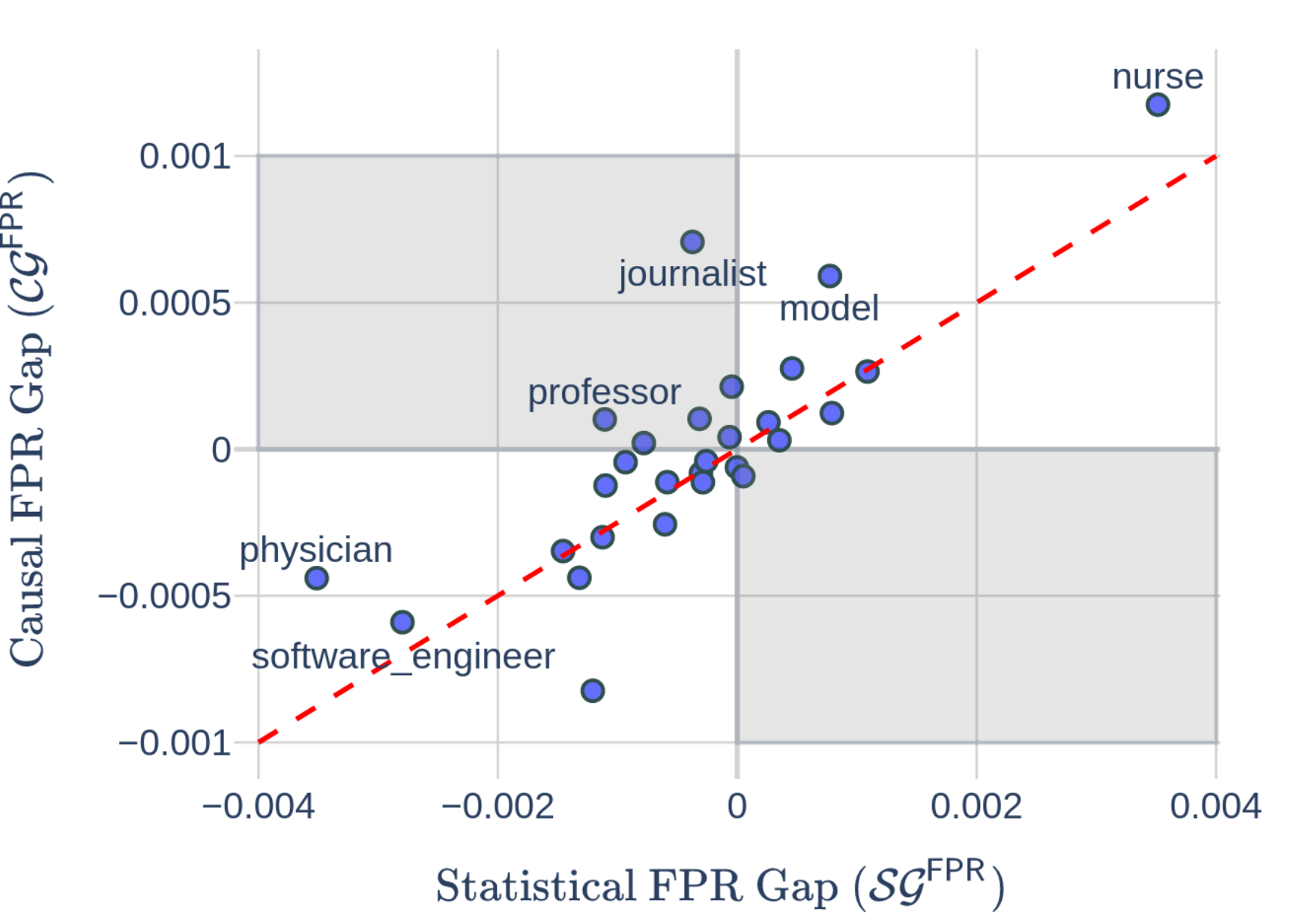}
         \caption{BERT-Base-Uncased}
         \label{fig:bert-baseline-fpr}
     \end{subfigure}
    \caption{Statistical and causal FPR gap on ALBERT-Large and BERT-Base-Uncased models with normal training. Red dashed line indicates $\mathcal{SP}=\mathcal{CP}$. Shaded areas represent $\mathcal{SP}$ and $\mathcal{CP}$ reporting opposite gender bias direction.}
    \label{fig:disparate-metrics-fpr}
\end{figure*}

\subsection{BoW Analysis}
\label{app:bow-analysis}

\begin{figure*}[!ht]
    \centering
    \includegraphics[width=\linewidth]{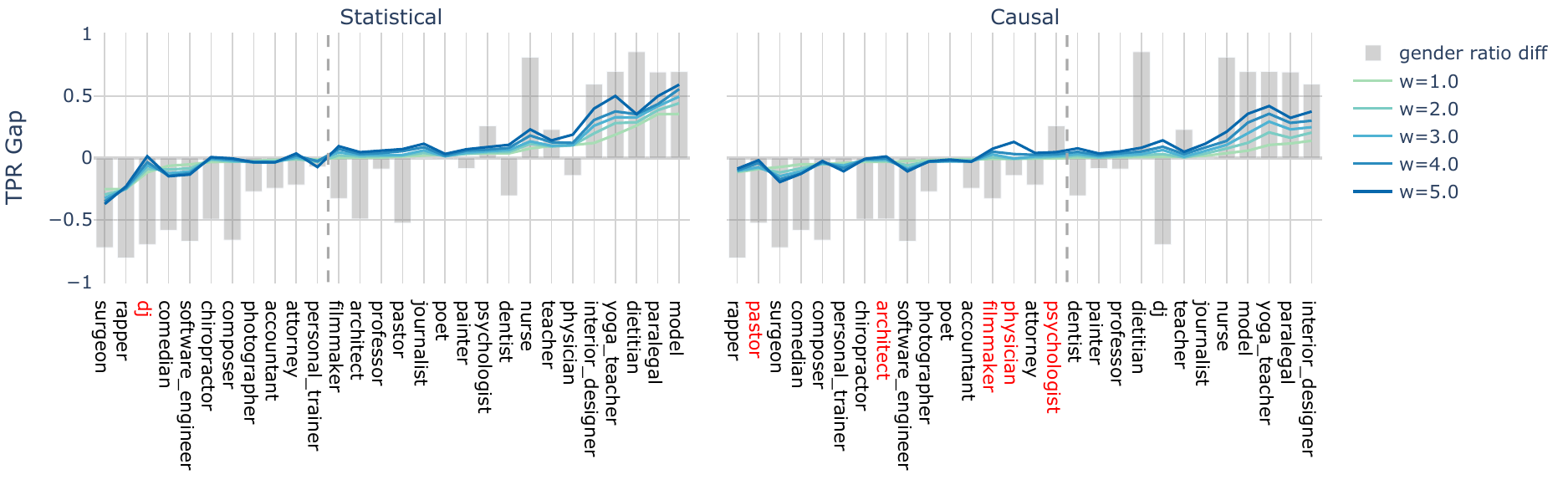}
    \caption{Statistical and causal TPR gaps of BoW model for each occupation when increasing the male token weights. Occupations are sorted by gap with $w=1$.}
    \label{fig:bow-increase-male-weight}
\end{figure*}

\begin{table}[!ht]
    \centering
    \begin{tabular}{c c c c c}
    \toprule
    Occupation & $\mathcal{SG}^{\mathsf{TPR}}$ & $\mathcal{CG}^{\mathsf{TPR}}$ & Diff & Gender ratio diff in train set \\
    \midrule
    \occupation{dj} & -0.115 & 0.008 & 0.123 & -0.695 \\
    \occupation{physician} & 0.105 & -0.005 & 0.110 & -0.140 \\
    \occupation{pastor} & 0.013 & -0.088 & 0.101 & -0.523 \\
    \occupation{psychologist} & 0.036 & -0.003 & 0.039 & 0.260 \\
    \occupation{poet} & 0.028 & -0.010 & 0.038 & -0.008 \\
    \occupation{architect} & 0.002 & -0.030 & 0.033 & -0.490 \\
    \occupation{filmmaker} & 0.02 & -0.009 & 0.011 & -0.325 \\
    \bottomrule
    \end{tabular}
    \caption{Occupations where statistical and causal TPR gap shows contradictory bias direction.}
    \label{tab:bow-bias-contradiction}
\end{table}

\clearpage

\begin{figure*}[!ht]
    \centering
    \includegraphics[width=0.8\linewidth]{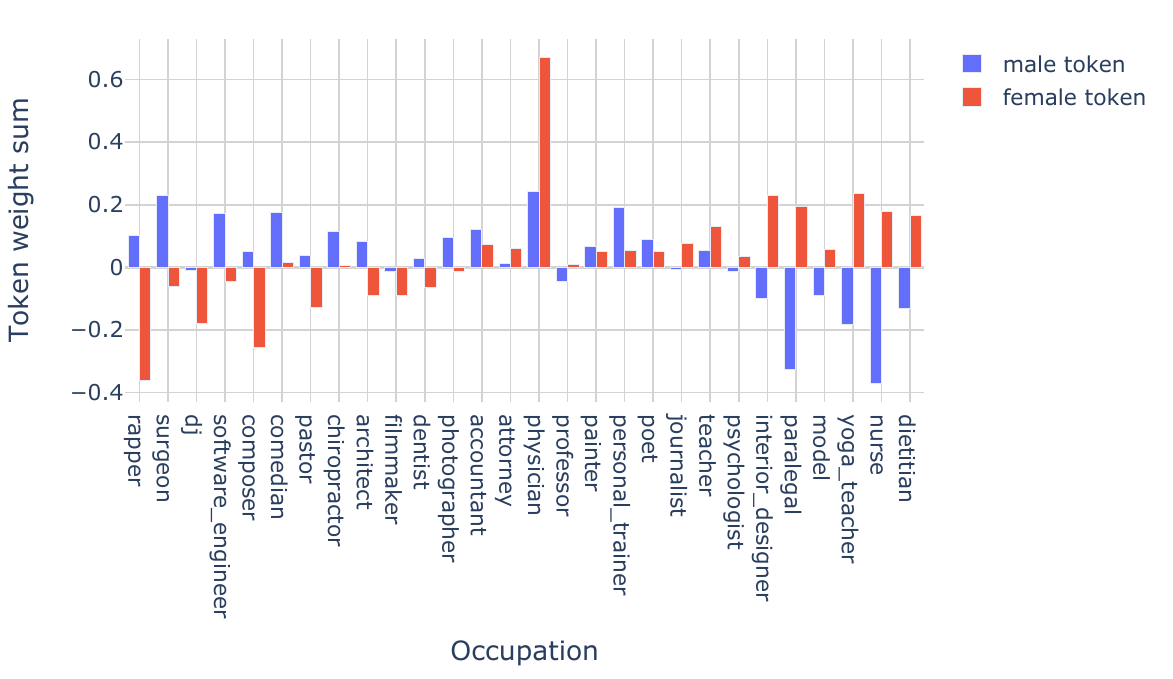}
    \caption{The sum of model weights for male and female gender tokens weighted by the token frequency in test examples of the occupation class.}
    \label{fig:bow-gender-token-weight-sum}
\end{figure*}

\begin{figure*}[!ht]
    \centering
    \includegraphics[width=\linewidth]{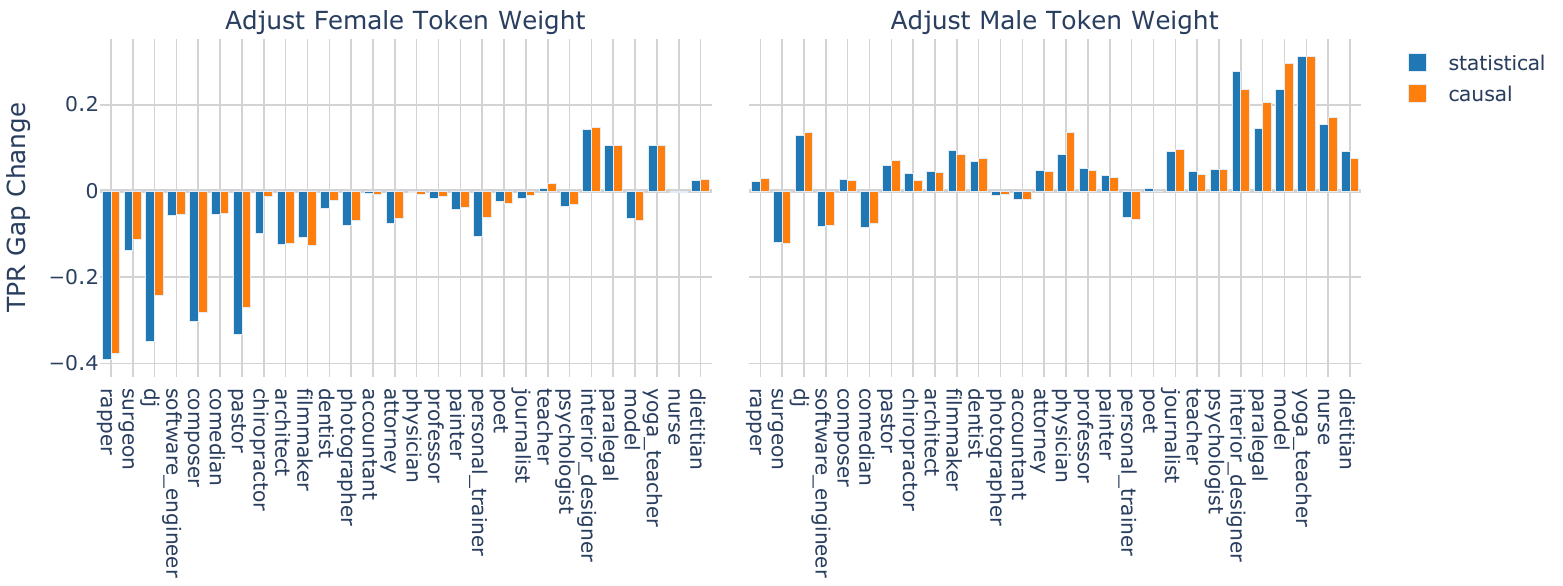}
    \caption{The TPR gap difference when increasing either female or male token weights from $w=1$ to $w=5$. Both metrics show similar patterns of TPR gap change for all occupations.}
    \label{fig:bow-tpr-gap-change}
\end{figure*}

%% file: appendix/train_details.tex
\section{Training Details}
\label{app:train-details}
\shortsection{Computing Infrastructure} All the models were trained on 4 Nvidia RTX 2080Ti GPUs.

\shortsection{BiasBios Dataset} We trained all the models with a learning rate of 2e-5 and batch size of 64. We fine-tuned the models for 5-8 epochs with early stopping and choose the model checkpoints with the best validation accuracy. Most models reach the best validation accuracy before epoch 5. We notice that ALBERT with subsampling requires training a few epochs longer than other models to reach comparable performance due to the downsized training data.

\shortsection{Jigsaw Dataset} We trained all the models with a learning rate of 1e-5 and batch size of 128 for 4 epochs with early stopping. Most models converge after 2-3 epochs.

%% file: appendix/biasbios_results.tex
\section{BiasBios Results}

\subsection{Overall Bias Scores}
\label{app:rms}

\input{tables/rms-biasbios-albert}
\input{tables/rms-biasbios-bert}
\clearpage

\subsection{Statistical vs Causal FPR Gap}
\label{app:sg-vs-cg-fpr}

\begin{figure*}[!ht]
     \centering
     \begin{subfigure}[b]{\linewidth}
         \centering
         \includegraphics[width=\textwidth]{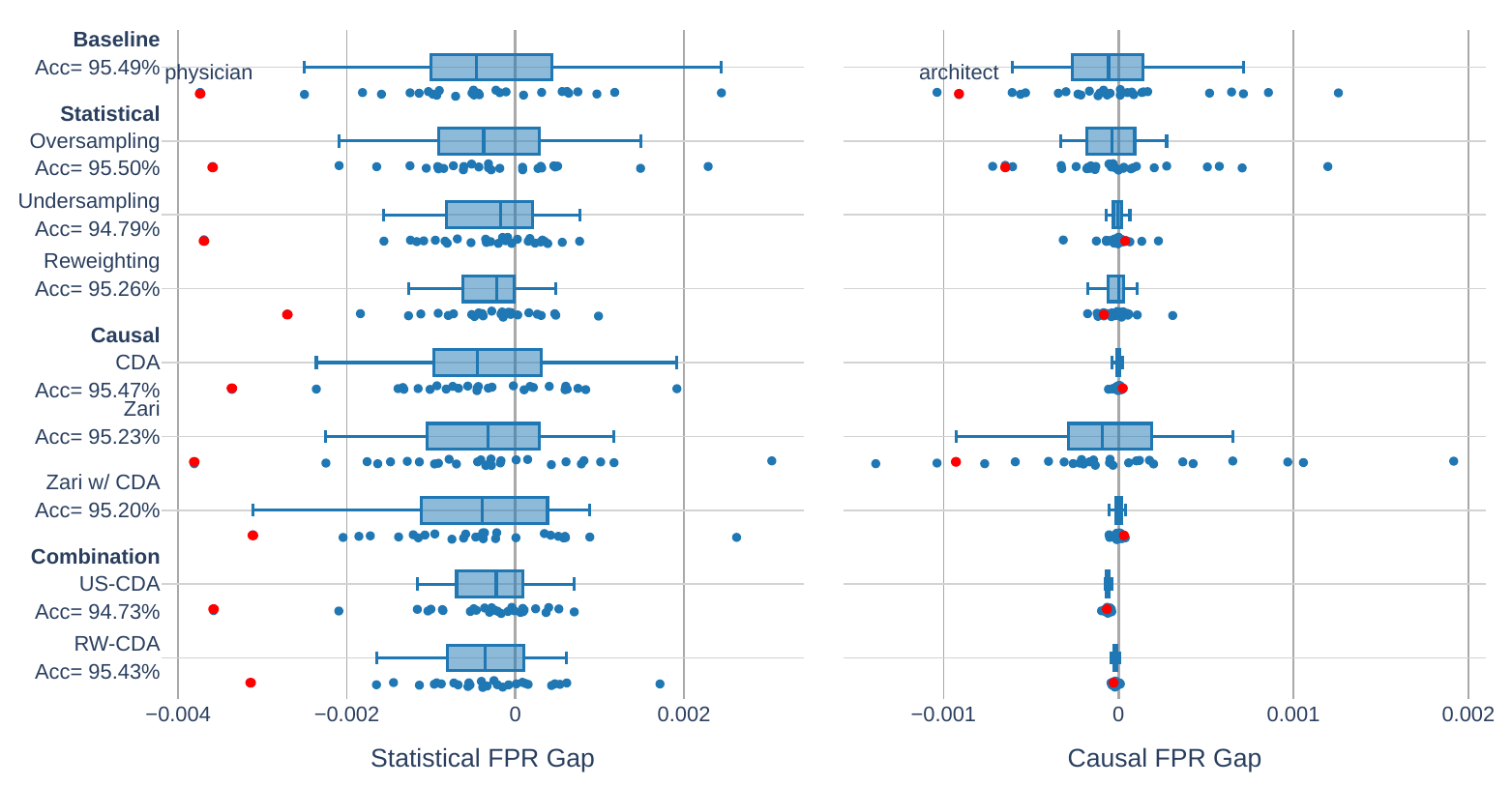}
         \caption{ALBERT-Large}
         \label{fig:albert-sg-vs-cg-fpr}
     \end{subfigure}
     \vfill
     \begin{subfigure}[b]{\linewidth}
         \centering
         \includegraphics[width=\textwidth]{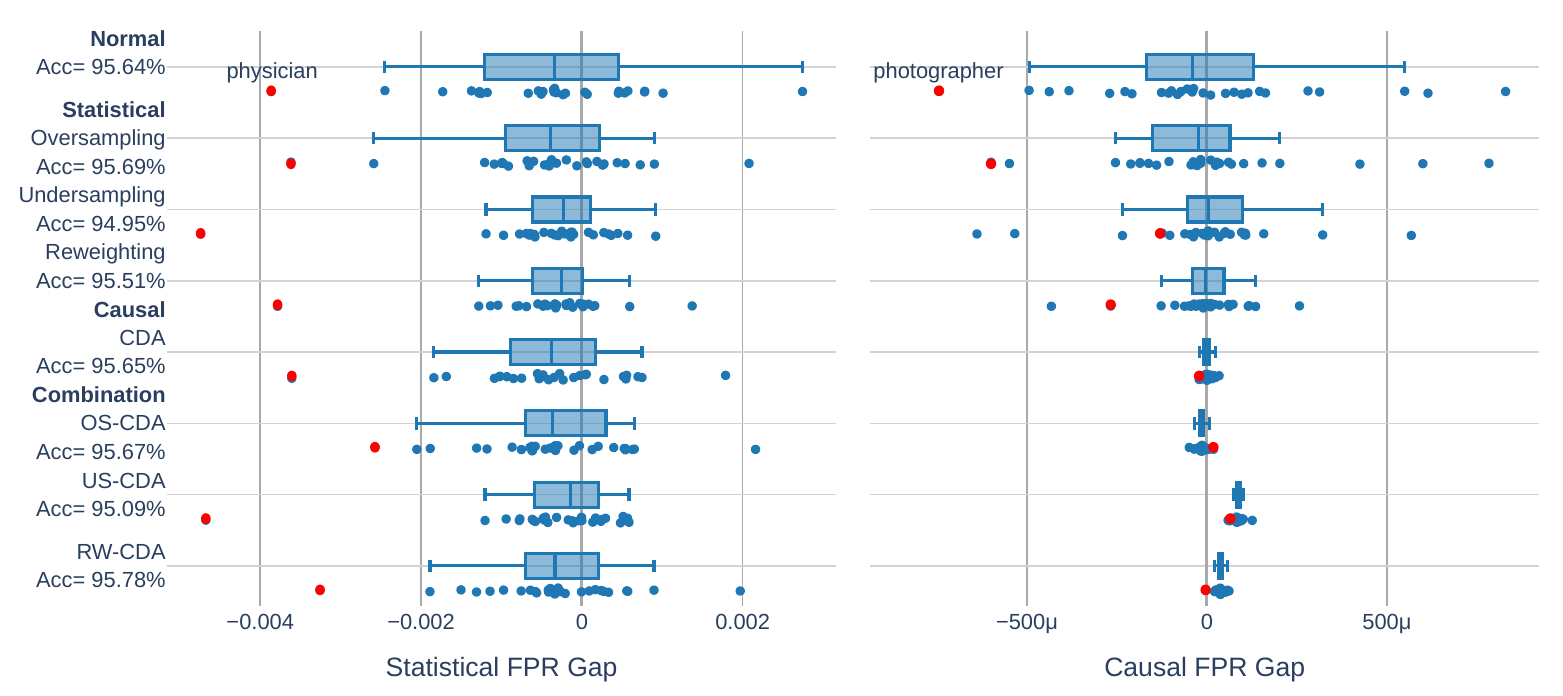}
         \caption{BERT-Base-Uncased}
         \label{fig:bert-sg-vs-cg-fpr}
     \end{subfigure}
     \caption{Statistical and Causal FPR gap per occupation, averaged over 3 different runs. Each data point is computed over test examples labeled with the same occupation. We show the outliers for normal training in red dots and how their values change with different debiasing methods. Causal-based debiasing methods perform particularly better on the causal FPR gap while statistical-based debiasing methods are able to reduce bias based on both metrics.}
     \label{fig:sg-vs-cg-fpr}
\end{figure*}

\clearpage
\subsection{Correlation to Gender Imbalances in Training Data}
\label{app:tpr-vs-train-female-ratio}

\begin{figure*}[htb]
     \centering
     \begin{subfigure}[b]{0.44\linewidth}
         \centering
         \includegraphics[width=\textwidth]{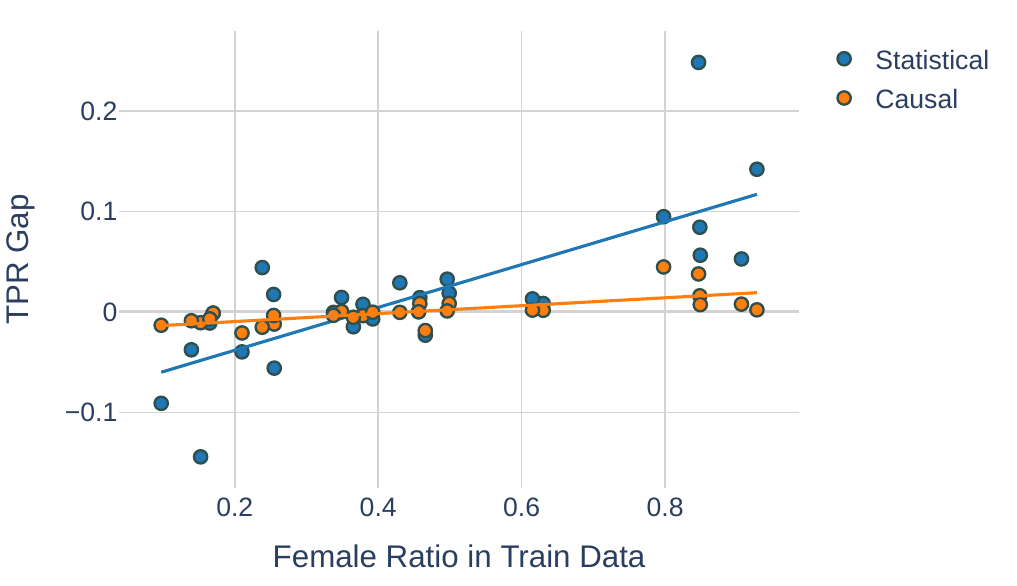}
         \caption{ALBERT Large}
     \end{subfigure}
     \hfill
     \begin{subfigure}[b]{0.54\linewidth}
         \centering
         \includegraphics[width=\textwidth]{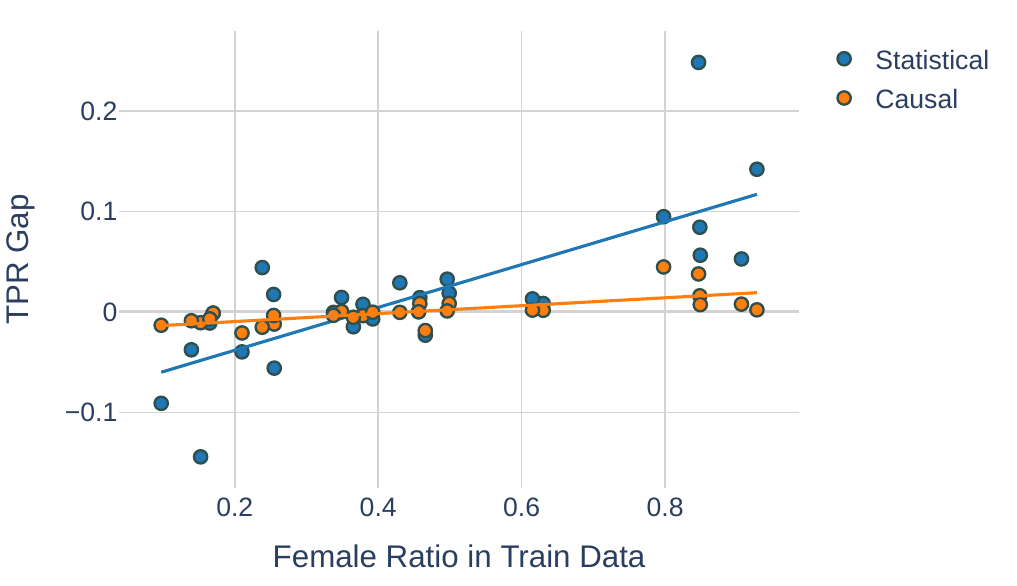}
         \caption{BERT Base Uncased}
     \end{subfigure}
     \caption{Statistical and causal TPR gap versus the female ratio of each occupation in the training data.}
     \label{fig:tpr-vs-train-female-ratio}
\end{figure*}

In \autoref{fig:tpr-vs-train-female-ratio}, we compare the statistical and causal TPR gap to the female ratio in the training data for each occupation. Both bias metrics show a positive correlation with the gender distribution in the training data. This observation is consistent with the results found in~\citet{de-arteaga2019bias}, where they measure the statistical TPR gap on non-transformer-based models such as BoW.

%% file: tables/rms-biasbios-albert.tex
\begin{table*}[htb]
    \centering
    \begin{filecontents*}{csv_files/rms_albert.csv}
method,Acc,statistical_tpr_gap,causal_tpr_gap,statistical_fpr_gap,causal_fpr_gap,Acc_std,statistical_tpr_gap_std,causal_tpr_gap_std,statistical_fpr_gap_std,causal_fpr_gap_std
Baseline,95.49280210970893,7.852732598985096,2.5691482873313447,0.1272525424854405,0.050787680279817546,0.12901425262710847,0.7610322369266873,0.5089458739808034,0.009463674893336917,0.004754275253875237
OS,95.50195412467755,6.4302539382618855,1.5904167628330843,0.11509836431699669,0.04124856197966413,0.042054705135791126,0.17162994497821113,0.03468871082799812,0.0035119801518512274,0.003028753116018101
US,94.7928424463675,5.60046457857962,0.5290792179158431,0.09698347902342264,0.010604967490735343,0.08132793708503994,0.4221712284406129,0.401582619788836,0.004500458691616204,0.0054262236232879985
RW,95.25518868404194,4.268675084051644,0.39138359234455894,0.08469967734713345,0.009657531539075191,0.05978935964811155,0.4267813180679897,0.09439333187502967,0.010864227967869592,0.0007606992423835676
CDA,95.46839673645925,7.265853066303212,0.20665351757556105,0.1125137314372404,0.003124893612605332,0.08977671285229023,0.8696036229249406,0.04328380046984562,0.007009948034388484,0.0001781691685271274
Zari,95.23180020134433,8.353409876591659,2.848913868908112,0.13245094010183917,0.06725598120150825,0.0904003690192014,0.5500137941420856,0.3409420856462831,0.005720874338100115,0.005484053317581562
Zari w/ CDA,95.1968869590566,7.559248679864044,0.21568719817571208,0.11881316636911977,0.003647641404674939,0.008219428475641038,0.787159669247259,0.04829781140713236,0.00794379231556425,0.0010485336507035292
OS-CDA,95.39382476264079,5.4029169217754385,0.12957799781242457,0.10932251066842755,0.012878057949784484,0.12839560411524184,0.1755509148626799,0.020028516789244544,0.005849031298010499,0.010548988646791916
US-CDA,94.72674456048296,4.9694758937500385,0.174201330184603,0.09646580878347438,0.007343352983982826,0.09326795378158835,0.2301171195174174,0.05056463801938985,0.01544796955716115,0.008844026841887936
RW-CDA,95.43416142120623,4.300479534666413,0.13723313113337227,0.09497009324817146,0.0075503477367051975,0.11455713254763487,0.4238208084269392,0.01967208236969756,0.010755828628587639,0.0040299852329243255
\end{filecontents*}
\pgfplotstabletypeset[
        col sep = comma,
        columns={method, acc, sp_tpr, sp_fpr, cp_tpr, cp_fpr},
        columns/method/.style={column name=Method, string type, column type/.add={}{|}},
        columns/acc/.style={string type, column name=Acc (\%), column type/.add={}{|}},
        columns/sp_tpr/.style={string type, column name=$\mathsf{TPR}$},
        columns/sp_fpr/.style={string type, column name=$\mathsf{FPR}$, column type/.add={}{|}},
        columns/cp_tpr/.style={string type, column name=$\mathsf{TPR}$},
        columns/cp_fpr/.style={string type, column name=$\mathsf{FPR}$, column type/.add={}{}},
        create on use/acc/.style={
            create col/assign/.code={%
              \edef\entry{\noexpand\pgfmathprintnumber[fixed, fixed zerofill, precision=2]{\thisrow{Acc}}$\pm$\noexpand\pgfmathprintnumber[fixed, fixed zerofill, precision=2]{\thisrow{Acc_std}}}%
              \pgfkeyslet{/pgfplots/table/create col/next content}\entry
            }
          },
        create on use/sp_tpr/.style={
            create col/assign/.code={%
              \edef\entry{\noexpand\pgfmathprintnumber[fixed, fixed zerofill, precision=3]{\thisrow{statistical_tpr_gap}}$\pm$\noexpand\pgfmathprintnumber[fixed, fixed zerofill, precision=3]{\thisrow{statistical_tpr_gap_std}}}%
              \pgfkeyslet{/pgfplots/table/create col/next content}\entry
            }
          },
        create on use/sp_fpr/.style={
            create col/assign/.code={%
              \edef\entry{\noexpand\pgfmathprintnumber[fixed, fixed zerofill, precision=3]{\thisrow{statistical_fpr_gap}}$\pm$\noexpand\pgfmathprintnumber[fixed, fixed zerofill, precision=3]{\thisrow{statistical_fpr_gap_std}}}%
              \pgfkeyslet{/pgfplots/table/create col/next content}\entry
            }
          },
        create on use/cp_tpr/.style={
            create col/assign/.code={%
              \edef\entry{\noexpand\pgfmathprintnumber[fixed, fixed zerofill, precision=3]{\thisrow{causal_tpr_gap}}$\pm$\noexpand\pgfmathprintnumber[fixed, fixed zerofill, precision=3]{\thisrow{causal_tpr_gap_std}}}%
              \pgfkeyslet{/pgfplots/table/create col/next content}\entry
            }
          },
        create on use/cp_fpr/.style={
            create col/assign/.code={%
              \edef\entry{\noexpand\pgfmathprintnumber[fixed, fixed zerofill, precision=3]{\thisrow{causal_fpr_gap}}$\pm$\noexpand\pgfmathprintnumber[fixed, fixed zerofill, precision=3]{\thisrow{causal_fpr_gap_std}}}%
              \pgfkeyslet{/pgfplots/table/create col/next content}\entry
            }
          },
        every head row/.style={before row={\toprule & & \multicolumn{2}{c|}{$\mathcal{SG}$} & \multicolumn{2}{c}{$\mathcal{CG}$}\\},after row=\midrule},
        every last row/.style={after row=\bottomrule},
        every row 1 column 1/.style={highlight bold},
        every row 3 column 2/.style={highlight bold},
        every row 3 column 3/.style={highlight bold},
        every row 7 column 4/.style={highlight bold},
        every row 4 column 5/.style={highlight bold},
        every row 0 column 0/.style={postproc cell content/.style=
        {@cell content=Normal}},
        every row 7 column 0/.style={postproc cell content/.style=
        {@cell content=\texttt{OS-CDA}}},
        every row 8 column 0/.style={postproc cell content/.style=
        {@cell content=\texttt{US-CDA}}},
        every row 9 column 0/.style={postproc cell content/.style=
        {@cell content=\texttt{RW-CDA}}},
    ]
    {csv_files/rms_albert.csv}%
    \caption{Root mean square of bias metrics for ALBERT-Large model fine-tuned with different debiasing methods. The results shown are averaged over 3 different runs. $\mathcal{SG}$ and $\mathcal{CG}$ are on a log scale with base $10^{-2}$.}
    \label{tab:rms-albert}
\end{table*}

%% file: tables/rms-biasbios-bert.tex
\begin{table*}[!h]
    \centering
    \begin{filecontents*}{csv_files/rms_bert.csv}
method,Acc,statistical_tpr_gap,causal_tpr_gap,statistical_fpr_gap,causal_fpr_gap,Acc_std,statistical_tpr_gap_std,causal_tpr_gap_std,statistical_fpr_gap_std,causal_fpr_gap_std
Baseline,95.64364087493263,7.472314568096851,1.4558265994445427,0.12939687123054858,0.0333783803495907,0.01883307657944469,0.8975926769333166,0.27148558528778133,0.004361410290809874,0.005136261497491685
Oversampling,95.68804509570636,6.161313928848555,0.8045536144458879,0.11633543951181712,0.02878532772646064,0.1681296632349022,0.2819700571008046,0.13411562739346622,0.01771115339102206,0.00789487382220812
Undersampling,94.95181633600774,5.25698266879856,0.5951212217460797,0.10807326613778129,0.02305896766852676,0.19270634930393635,0.8651178280645653,0.08310525533392095,0.017194114734572892,0.00044952167435606224
Reweighting,95.50703857743791,4.629992278769771,0.37666589820953705,0.09586673538975322,0.013742650753928254,0.059697047544970426,0.2876222044940471,0.07407992909337022,0.007874356291413935,0.0038309344829822786
CDA,95.64601361955413,6.489813742268917,0.13752568586278602,0.1087146315350301,0.0021757503926586193,0.07959293576623182,1.1594915244179154,0.04634716313603181,0.010563780257856639,0.0007599380096624305
Oversampling w/ CDA,95.66567350356081,5.484522549682014,0.12059039796861247,0.10632334089407944,0.00472233599002671,0.09216380014092444,0.32736974327682106,0.03304400064975026,0.021531115892648466,0.0033771173386510627
Undersampling w/ CDA,95.08977448757192,4.672822884693684,0.1313721075233039,0.10414511433551155,0.008742377097301026,0.11573060371137665,0.2695839504284957,0.011661669065807269,0.007249500180201125,0.002136349388306802
Reweight w/ CDA,95.77719250077115,4.600707378815871,0.14785513170804224,0.1020121424952245,0.004223775650531647,0.0696576213606033,0.19049277538047343,0.020649140591329742,0.0024552139743957667,0.0025007720643379404
\end{filecontents*}
\pgfplotstabletypeset[
        col sep = comma,
        columns={method, acc, sp_tpr, sp_fpr, cp_tpr, cp_fpr},
        columns/method/.style={column name=Method, string type, column type/.add={}{|}},
        columns/acc/.style={string type, column name=Acc (\%), column type/.add={}{|}},
        columns/sp_tpr/.style={string type, column name=$\mathsf{TPR}$},
        columns/sp_fpr/.style={string type, column name=$\mathsf{FPR}$, column type/.add={}{|}},
        columns/cp_tpr/.style={string type, column name=$\mathsf{TPR}$},
        columns/cp_fpr/.style={string type, column name=$\mathsf{FPR}$, column type/.add={}{}},
        create on use/acc/.style={
            create col/assign/.code={%
              \edef\entry{\noexpand\pgfmathprintnumber[fixed, fixed zerofill, precision=2]{\thisrow{Acc}}$\pm$\noexpand\pgfmathprintnumber[fixed, fixed zerofill, precision=2]{\thisrow{Acc_std}}}%
              \pgfkeyslet{/pgfplots/table/create col/next content}\entry
            }
          },
        create on use/sp_tpr/.style={
            create col/assign/.code={%
              \edef\entry{\noexpand\pgfmathprintnumber[fixed, fixed zerofill, precision=3]{\thisrow{statistical_tpr_gap}}$\pm$\noexpand\pgfmathprintnumber[fixed, fixed zerofill, precision=3]{\thisrow{statistical_tpr_gap_std}}}%
              \pgfkeyslet{/pgfplots/table/create col/next content}\entry
            }
          },
        create on use/sp_fpr/.style={
            create col/assign/.code={%
              \edef\entry{\noexpand\pgfmathprintnumber[fixed, fixed zerofill, precision=3]{\thisrow{statistical_fpr_gap}}$\pm$\noexpand\pgfmathprintnumber[fixed, fixed zerofill, precision=3]{\thisrow{statistical_fpr_gap_std}}}%
              \pgfkeyslet{/pgfplots/table/create col/next content}\entry
            }
          },
        create on use/cp_tpr/.style={
            create col/assign/.code={%
              \edef\entry{\noexpand\pgfmathprintnumber[fixed, fixed zerofill, precision=3]{\thisrow{causal_tpr_gap}}$\pm$\noexpand\pgfmathprintnumber[fixed, fixed zerofill, precision=3]{\thisrow{causal_tpr_gap_std}}}%
              \pgfkeyslet{/pgfplots/table/create col/next content}\entry
            }
          },
        create on use/cp_fpr/.style={
            create col/assign/.code={%
              \edef\entry{\noexpand\pgfmathprintnumber[fixed, fixed zerofill, precision=3]{\thisrow{causal_fpr_gap}}$\pm$\noexpand\pgfmathprintnumber[fixed, fixed zerofill, precision=3]{\thisrow{causal_fpr_gap_std}}}%
              \pgfkeyslet{/pgfplots/table/create col/next content}\entry
            }
          },
        every head row/.style={before row={\toprule & & \multicolumn{2}{c|}{$\mathcal{SG}$} & \multicolumn{2}{c}{$\mathcal{CG}$}\\},after row=\midrule},
        every last row/.style={after row=\bottomrule},
        every row 7 column 1/.style={highlight bold},
        every row 7 column 2/.style={highlight bold},
        every row 3 column 3/.style={highlight bold},
        every row 5 column 4/.style={highlight bold},
        every row 4 column 5/.style={highlight bold},
        every row 1 column 0/.style={postproc cell content/.style=
        {@cell content=OS}},
        every row 2 column 0/.style={postproc cell content/.style=
        {@cell content=US}},
        every row 3 column 0/.style={postproc cell content/.style=
        {@cell content=RW}},
        every row 5 column 0/.style={postproc cell content/.style=
        {@cell content=\texttt{OS-CDA}}},
        every row 6 column 0/.style={postproc cell content/.style=
        {@cell content=\texttt{US-CDA}}},
        every row 7 column 0/.style={postproc cell content/.style=
        {@cell content=\texttt{RW-CDA}}}
    ]
    {csv_files/rms_bert.csv}%
    \caption{Root mean square of bias metrics for BERT-Base-Uncased model fine-tuned with different debiasing methods. The values shown are averaged over 3 different runs on a log scale with base $10^{-2}$.}
    \label{tab:rms-bert}
\end{table*}

%% file: appendix/jigsaw_results.tex
\section{Jigsaw Results}

\subsection{Overall Bias Scores for ALBERT Model}
\label{app:jigsaw-bias-score-albert}
\input{tables/jigsaw-ppr-gap-albert}

\subsection{Combination Strategies Comparison}
\label{app:combination-strategy-comparison}

\autoref{tab:jigsaw-alternative-comparison-resampling} shows the performance of two different strategies of combining resampling and CDA. Resample $\rightarrow$ CDA performs resampling first, then applies CDA on the resampled set. CDA $\rightarrow$ Resample performs CDA first, then resamples the original and the counterfactual sets separately. The original examples are resampled based on the original gender distribution. The counterfactual examples are resampled based on their counterfactual genders (not the gender of the original example they originated from). The difference between the two methods is that Resample $\rightarrow$ CDA uses the original gender label for both original and counterfactual examples while CDA $\rightarrow$ Resample considers the counterfactual gender for the counterfactual examples during resampling. We find that the second method performs better on $\mathcal{SG}^{\mathsf{PPR}}$ but increases $\mathcal{CG}^{\mathsf{PPR}}$ compared to the first method. The increase in the causal bias metric may be due to separate resampling on original and counterfactual sets, meaning that some of them may not come in pairs. Nonetheless, the performance still exceeds CDA.

\input{tables/jigsaw-resampling-strategy-comparison}

\autoref{tab:jigsaw-alternative-comparison-reweight} shows the performance of using different reweighting strategies on counterfactual examples for \texttt{RW-CDA}. We tried \texttt{RW-CDA} method for training on BiasBios dataset, which uses the same weight for both the original and counterfactual examples (first row in \autoref{tab:jigsaw-alternative-comparison-reweight}). It is not effective at reducing $\mathcal{SG}^{\mathsf{PPR}}$, but very effective on $\mathcal{CG}^{\mathsf{PPR}}$. We think it may be due to the gender signals of some examples being flipped by CDA. We then tried using weights that correspond to the counterfactual gender for the counterfactual examples. This decreases bias on $\mathcal{SG}^{\mathsf{PPR}}$, but increases bias on $\mathcal{CG}^{\mathsf{PPR}}$. We found that setting the weight to 1 for all counterfactual examples gives the best overall balance between $\mathcal{SG}^{\mathsf{PPR}}$ and $\mathcal{CG}^{\mathsf{PPR}}$. It also outperforms other strategies on $\mathcal{SG}^{\mathsf{PPR}}$.

\input{tables/jigsaw-reweight-strategy-comparison}

\subsection{General Performance}
\label{app:jigsaw-auc}

\begin{table*}[!h]
    \centering
    \begin{filecontents*}{csv_files/jigsaw-auc.csv}
method,auc,auc_std, auc_bert, auc_std_bert
Normal,0.9303964757709251,0.0016392888077654928,0.9250367107195302,0.003485778992795655
CDA,0.9296255506607929,0.0017397937754879188,0.9279735682819383,0.0024914654186427606
Zari w/ CDA,0.928157121879589,0.004947918661582762,0.0,0.0
OS,0.9310939794419971,0.0012263798156013426,0.9317180616740088,0.0016293936138892851
US,0.9287077826725405,0.002972435019089172,0.9244860499265786,0.003942943195917029
RW,0.9301027900146842,0.004907991179550013,0.9290014684287812,0.002683651524595759
OS-CDA,0.9295888399412628,0.0025817842837763992,0.9310205580029368,0.0016597143253241394
US-CDA,0.9289280469897211,0.002591683200399832,0.9305433186490456,0.0022468639865050343
RW-CDA,0.9291483113069017,0.0019905200751433034,0.9303964757709251,0.0026457199334401337
\end{filecontents*}
\pgfplotstabletypeset[
        col sep = comma,
        columns={method, albert-auc, bert-auc},
        columns/method/.style={column name=Method, string type, column type/.add={}{|}},
        columns/albert-auc/.style={string type, column name=AUC (ALBERT), column type/.add={}{|}},
        columns/bert-auc/.style={string type, column name=AUC (BERT)},
        create on use/albert-auc/.style={
            create col/assign/.code={%
              \edef\entry{\noexpand\pgfmathprintnumber[fixed, fixed zerofill, precision=3]{\thisrow{auc}}$\pm$\noexpand\pgfmathprintnumber[fixed, fixed zerofill, precision=3]{\thisrow{auc_std}}}%
              \pgfkeyslet{/pgfplots/table/create col/next content}\entry
            }
          },
        create on use/bert-auc/.style={
            create col/assign/.code={%
              \edef\entry{\noexpand\pgfmathprintnumber[fixed, fixed zerofill, precision=3]{\thisrow{auc_bert}}$\pm$\noexpand\pgfmathprintnumber[fixed, fixed zerofill, precision=3]{\thisrow{auc_std_bert}}}%
              \pgfkeyslet{/pgfplots/table/create col/next content}\entry
            }
        },
        every head row/.style={before row=\toprule, after row=\midrule},
        every last row/.style={after row=\bottomrule},
        every row 2 column 2/.style={postproc cell content/.style=
        {@cell content=---}},
        every row 6 column 0/.style={postproc cell content/.style=
        {@cell content=\texttt{OS-CDA}}},
        every row 7 column 0/.style={postproc cell content/.style=
        {@cell content=\texttt{US-CDA}}},
        every row 8 column 0/.style={postproc cell content/.style=
        {@cell content=\texttt{RW-CDA}}}
    ]
    {csv_files/jigsaw-auc.csv}
    \caption{AUC scores of different debiasing methods. The results shown are averaged over 5 different runs.}
    \label{tab:jigsaw-auc}
\end{table*}

\subsection{Gender Label Annotation Agreement}
\label{sec:gender-agreement}
\begin{figure*}[htb]
    \centering
    \begin{subfigure}[b]{\linewidth}
        \includegraphics[width=\linewidth]{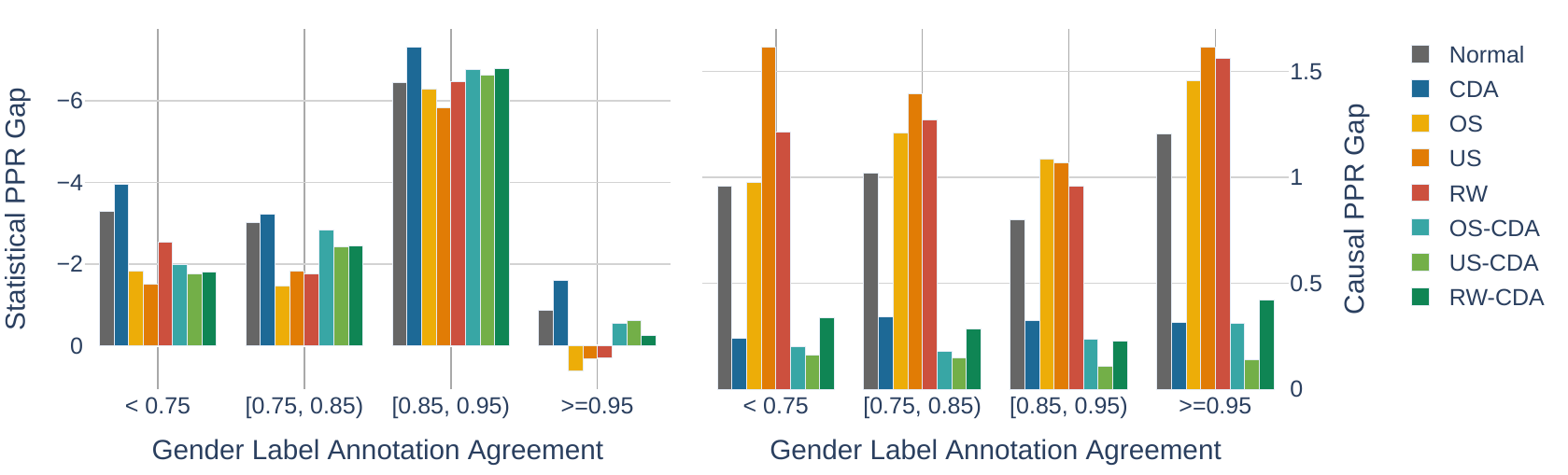}
        \caption{BERT-Base-Uncased}
    \end{subfigure}
    \hfill
    \begin{subfigure}[b]{\linewidth}
        \includegraphics[width=\linewidth]{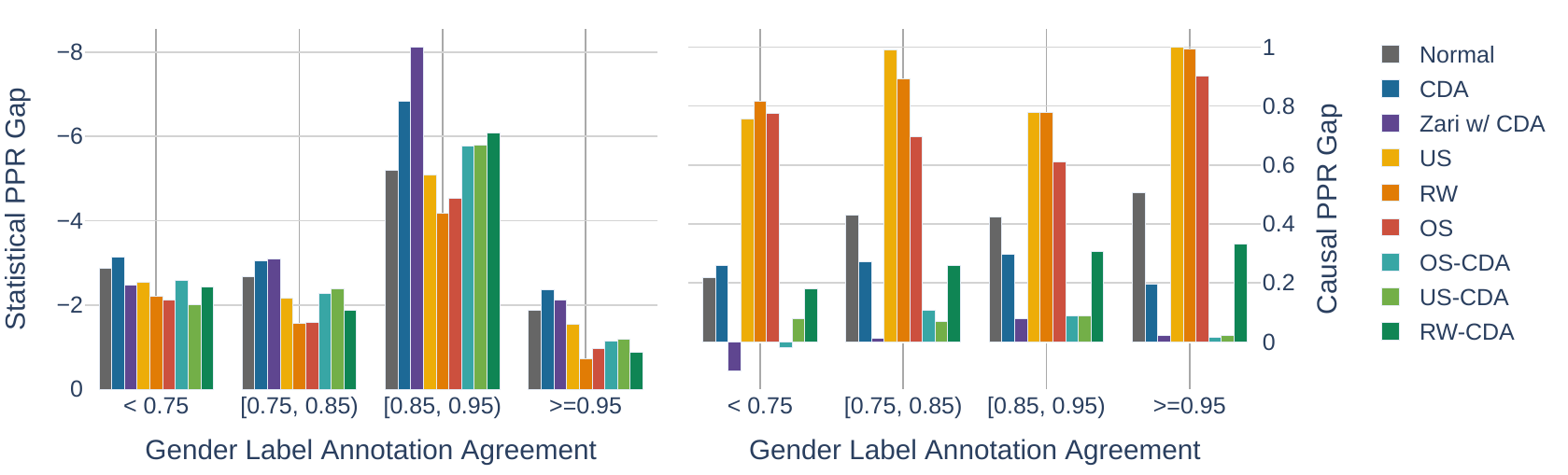}
        \caption{ALBERT-Large}
    \end{subfigure}
    \caption{Statistical and Causal PPR Gap of examples with different range of gender label annotation agreement.}
    \label{fig:jigsaw-ppr-vs-gender-agreement}
\end{figure*}

We test if gender label annotation agreement in the Jigsaw dataset has an effect on the bias scores. In \autoref{fig:jigsaw-ppr-vs-gender-agreement}, we show statistical and causal PPR gap of examples with different range of annotation agreement for each debiasing methods. All methods have the highest score of statistical PPR gap at [0.85, 0.96) including the normal training method and have the lowest score when annotation agreement >=0.95. On the other hand, causal PPR gap of each debiasing method remain similar at different range of gender annotation agreement.

%% file: tables/jigsaw-ppr-gap-albert.tex
\begin{table*}[!h]
    \centering
    \resizebox{\linewidth}{!}{%
    \begin{filecontents*}{csv_files/jigsaw-ppr-gap-albert.csv}
method,auc,auc_std,statistical_ppr_gap,statistical_ppr_gap_std,causal_ppr_gap,causal_ppr_gap_std,statistical_tpr_gap,statistical_tpr_gap_std,causal_tpr_gap,causal_tpr_gap_std,statistical_tpr_neg_gap,statistical_tpr_neg_gap_std,causal_tpr_neg_gap,causal_tpr_neg_gap_std
Normal,0.9303964757709251,0.0016392888077654928,-2.733074685973254,0.4195197629798805,0.4221732745961817,0.20636518829009087,-4.601146220570013,3.646519310890394,1.9030520646319569,1.365397008390057,1.2115453695401213,0.44860093158261727,-0.2535268861173584,0.08033891925736454
CDA,0.9296255506607929,0.0017397937754879188,-3.141796225418124,0.5928899271022964,0.20190895741556553,0.0795869434239309,-3.560512185047502,3.0786373602068955,0.8617594254937163,0.6659690122438566,1.660400950962888,0.3613739470539493,-0.1267634430586792,0.06768736600796095
Zari w/ CDA,0.928157121879589,0.004947918661582762,-2.8891410077614843,0.9798593391067275,-0.04772393538913386,0.11643723588218621,-5.6786968194960785,2.0972870412664286,-0.3231597845601436,0.5722577181511392,1.3147480140286238,0.9207328466285792,0.01635657329789409,0.07245980432087241
US,0.9287077826725405,0.002972435019089172,-2.3672824620439936,0.5839852142133551,0.998531571218796,0.09891478403309438,-2.572542337876911,2.748460595483535,4.201077199281868,0.8156421322513856,1.0251574655248508,0.45401785412404766,-0.6338172152933961,0.0827988416737542
RW,0.9301027900146842,0.004907991179550013,-1.7046425114158708,0.21456310592735778,0.9471365638766519,0.2467004055273344,-2.068360181743082,2.1532016268655587,4.129263913824058,0.300416526583151,0.3869970632028408,0.2929095899155954,-0.5847474953997137,0.2775200081939424
OS,0.9310939794419971,0.0012263798156013426,-1.7863923234160712,0.23619212564590567,0.8149779735682816,0.21705940527304673,-3.1833436596447737,2.748862044939398,3.985637342908438,0.798065018703514,0.4764514660677954,0.22292479646804295,-0.453894909016561,0.2078636684054604
OS-CDA,0.9295888399412628,0.0025817842837763992,-2.2944685746209563,0.41972248818269986,0.01468428781204123,0.10728075768136922,-3.3952912019826553,2.7419060594340747,0.2872531418312388,0.6869345231493355,0.8347455187107689,0.30122411234288954,0.016356573297894094,0.05981900976621465
US-CDA,0.9289280469897211,0.002591683200399832,-2.2191448124783704,0.23167370454078703,0.08443465491923624,0.09615125443251644,-2.5684118133002865,2.5980499119249196,0.3590664272890485,0.2538983056325126,0.8845696125127867,0.30391820416758347,-0.05315886321815579,0.10552842282203181
RW-CDA,0.9291483113069017,0.0019905200751433034,-1.9608203602578402,0.24507246871135197,0.2386196769456679,0.09066878880490817,-1.9767141676992985,1.3602709782373827,0.9694793536804309,0.7323546877691612,0.7648697636029111,0.2502918004742698,-0.1553874463299939,0.06915368851068396
\end{filecontents*}
\pgfplotstabletypeset[
        col sep = comma,
        columns={method, sg-ppr, cg-ppr, sg-tpr, cg-tpr, sg-tpr-2, cg-tpr-2},
        columns/method/.style={column name=Method, string type, column type/.add={}{|}},
        columns/sg-ppr/.style={string type, column name=$\mathcal{SG}^{\mathsf{PPR}}$},
        columns/cg-ppr/.style={string type, column name=$\mathcal{CG}^{\mathsf{PPR}}$},
        columns/sg-tpr/.style={string type, column name=$\mathcal{SG}^{\mathsf{TPR}}_{y=1}$},
        columns/cg-tpr/.style={string type, column name=$\mathcal{CG}^{\mathsf{TPR}}_{y=1}$},
        columns/sg-tpr-2/.style={string type, column name=$\mathcal{SG}^{\mathsf{TPR}}_{y=0}$},
        columns/cg-tpr-2/.style={string type, column name=$\mathcal{CG}^{\mathsf{TPR}}_{y=0}$},
        create on use/auc-mixed/.style={
            create col/assign/.code={%
              \edef\entry{\noexpand\pgfmathprintnumber[fixed, fixed zerofill, precision=3]{\thisrow{auc}}$\pm$\noexpand\pgfmathprintnumber[fixed, fixed zerofill, precision=3]{\thisrow{auc_std}}}%
              \pgfkeyslet{/pgfplots/table/create col/next content}\entry
            }
          },
        create on use/sg-ppr/.style={
            create col/assign/.code={%
              \edef\entry{\noexpand\pgfmathprintnumber[fixed, fixed zerofill, precision=2]{\thisrow{statistical_ppr_gap}}$\pm$\noexpand\pgfmathprintnumber[fixed, fixed zerofill, precision=2]{\thisrow{statistical_ppr_gap_std}}}%
              \pgfkeyslet{/pgfplots/table/create col/next content}\entry
            }
          },
        create on use/cg-ppr/.style={
            create col/assign/.code={%
              \edef\entry{\noexpand\pgfmathprintnumber[fixed, fixed zerofill, precision=2]{\thisrow{causal_ppr_gap}}$\pm$\noexpand\pgfmathprintnumber[fixed, fixed zerofill, precision=2]{\thisrow{causal_ppr_gap_std}}}%
              \pgfkeyslet{/pgfplots/table/create col/next content}\entry
            }
          },
        create on use/sg-tpr/.style={
            create col/assign/.code={%
              \edef\entry{\noexpand\pgfmathprintnumber[fixed, fixed zerofill, precision=2]{\thisrow{statistical_tpr_gap}}$\pm$\noexpand\pgfmathprintnumber[fixed, fixed zerofill, precision=2]{\thisrow{statistical_tpr_gap_std}}}%
              \pgfkeyslet{/pgfplots/table/create col/next content}\entry
            }
          },
        create on use/cg-tpr/.style={
            create col/assign/.code={%
              \edef\entry{\noexpand\pgfmathprintnumber[fixed, fixed zerofill, precision=2]{\thisrow{causal_tpr_gap}}$\pm$\noexpand\pgfmathprintnumber[fixed, fixed zerofill, precision=2]{\thisrow{causal_tpr_gap_std}}}%
              \pgfkeyslet{/pgfplots/table/create col/next content}\entry
            }
          },
        create on use/sg-tpr-2/.style={
            create col/assign/.code={%
              \edef\entry{\noexpand\pgfmathprintnumber[fixed, fixed zerofill, precision=2]{\thisrow{statistical_tpr_neg_gap}}$\pm$\noexpand\pgfmathprintnumber[fixed, fixed zerofill, precision=2]{\thisrow{statistical_tpr_neg_gap_std}}}%
              \pgfkeyslet{/pgfplots/table/create col/next content}\entry
            }
          },
        create on use/cg-tpr-2/.style={
            create col/assign/.code={%
              \edef\entry{\noexpand\pgfmathprintnumber[fixed, fixed zerofill, precision=2]{\thisrow{causal_tpr_neg_gap}}$\pm$\noexpand\pgfmathprintnumber[fixed, fixed zerofill, precision=2]{\thisrow{causal_tpr_neg_gap_std}}}%
              \pgfkeyslet{/pgfplots/table/create col/next content}\entry
            }
          },
        every head row/.style={before row={\toprule},after row=\midrule},
        every last row/.style={after row=\bottomrule},
        every row 4 column 1/.style={highlight bold},
        every row 6 column 2/.style={highlight bold},
        every row 8 column 3/.style={highlight bold},
        every row 6 column 4/.style={highlight bold},
        every row 4 column 5/.style={highlight bold},
        every row 6 column 6/.style={highlight bold},
        every row 6 column 0/.style={postproc cell content/.style=
        {@cell content=\texttt{OS-CDA}}},
        every row 7 column 0/.style={postproc cell content/.style=
        {@cell content=\texttt{US-CDA}}},
        every row 8 column 0/.style={postproc cell content/.style=
        {@cell content=\texttt{RW-CDA}}}
    ]
    {csv_files/jigsaw-ppr-gap-albert.csv}%
    }
    \caption{Bias evaluation results evaluated on the Jigsaw dataset with ALBERT-Large model. The results shown are averaged over 5 different runs. All values are on a log scale with base $10^{-2}$.}
    \label{tab:jigsaw-albert}
\end{table*}

%% file: tables/jigsaw-resampling-strategy-comparison.tex
\begin{table*}[!h]
    \centering
    \begin{filecontents*}{csv_files/jigsaw-alternative-comparison-resampling.csv}
strategy,method,statistical_ppr_gap_bert,statistical_ppr_gap_std_bert,causal_ppr_gap_bert,causal_ppr_gap_std_bert,statistical_ppr_gap_albert,statistical_ppr_gap_std_albert,causal_ppr_gap_albert,causal_ppr_gap_std_albert
v1,OS-CDA,-2.726827519730519,0.719007057577723,0.011013215859030923,0.08578062735350887,-2.514853188153227,0.49159830130055,0.0036710719530105153,0.0815936518755245
v1,US-CDA,-2.1218293739279885,0.5085817994689734,0.11747430249632873,0.11409890064552211,-2.8767292369444206,0.7809274520820552,0.022026431718061568,0.1340821025532283
v2,OS-CDA,-2.089619315027118,0.3030007486929356,0.17621145374449365,0.1598497874968964,-2.2944685746209563,0.41972248818269986,0.01468428781204123,0.10728075768136922
v2,US-CDA,-1.9004842929865922,0.18717726853174252,0.11380323054331898,0.11338799714743356,-2.2191448124783704,0.23167370454078703,0.08443465491923624,0.09615125443251644
\end{filecontents*}
\pgfplotstabletypeset[
        col sep = comma,
        columns={strategy, method, sg-ppr-bert, cg-ppr-bert, sg-ppr-albert, cg-ppr-albert},
        columns/strategy/.style={column name=Strategy, string type, column type/.add={}{|},
            assign cell content/.code={
                \ifnum\pgfplotstablerow=0
                    \pgfkeyssetvalue{/pgfplots/table/@cell content}{\multirow{2}{*}{Resample $\rightarrow$ CDA}}%
                \else
                    \ifnum\pgfplotstablerow=2
                        \pgfkeyssetvalue{/pgfplots/table/@cell content}{\multirow{2}{*}{CDA $\rightarrow$ Resample}}
                    \else
                        \pgfkeyssetvalue{/pgfplots/table/@cell content}{}
                    \fi 
                \fi
            },
        },
        columns/method/.style={column name=Method, string type, column type/.add={}{|}},
        columns/sg-ppr-bert/.style={string type, column name=$\mathcal{SG}^{\mathsf{PPR}}$},
        columns/cg-ppr-bert/.style={string type, column name=$\mathcal{CG}^{\mathsf{PPR}}$, column type/.add={}{|}},
        columns/sg-ppr-albert/.style={string type, column name=$\mathcal{SG}^{\mathsf{PPR}}$},
        columns/cg-ppr-albert/.style={string type, column name=$\mathcal{CG}^{\mathsf{PPR}}$},
        create on use/sg-ppr-bert/.style={
            create col/assign/.code={%
              \edef\entry{\noexpand\pgfmathprintnumber[fixed, fixed zerofill, precision=2]{\thisrow{statistical_ppr_gap_bert}}$\pm$\noexpand\pgfmathprintnumber[fixed, fixed zerofill, precision=2]{\thisrow{statistical_ppr_gap_std_bert}}}%
              \pgfkeyslet{/pgfplots/table/create col/next content}\entry
            }
          },
        create on use/cg-ppr-bert/.style={
            create col/assign/.code={%
              \edef\entry{\noexpand\pgfmathprintnumber[fixed, fixed zerofill, precision=3]{\thisrow{causal_ppr_gap_bert}}$\pm$\noexpand\pgfmathprintnumber[fixed, fixed zerofill, precision=3]{\thisrow{causal_ppr_gap_std_bert}}}%
              \pgfkeyslet{/pgfplots/table/create col/next content}\entry
            }
          },
        create on use/sg-ppr-albert/.style={
            create col/assign/.code={%
              \edef\entry{\noexpand\pgfmathprintnumber[fixed, fixed zerofill, precision=2]{\thisrow{statistical_ppr_gap_albert}}$\pm$\noexpand\pgfmathprintnumber[fixed, fixed zerofill, precision=2]{\thisrow{statistical_ppr_gap_std_albert}}}%
              \pgfkeyslet{/pgfplots/table/create col/next content}\entry
            }
          },
        create on use/cg-ppr-albert/.style={
            create col/assign/.code={%
              \edef\entry{\noexpand\pgfmathprintnumber[fixed, fixed zerofill, precision=3]{\thisrow{causal_ppr_gap_albert}}$\pm$\noexpand\pgfmathprintnumber[fixed, fixed zerofill, precision=3]{\thisrow{causal_ppr_gap_std_albert}}}%
              \pgfkeyslet{/pgfplots/table/create col/next content}\entry
            }
          },
        every head row/.style={before row={\toprule & & \multicolumn{2}{c|}{BERT-Base-Uncased} & \multicolumn{2}{c}{ALBERT-Large}\\},after row=\midrule},
        every row no 1/.style={after row=\midrule},
        every last row/.style={after row=\bottomrule},
        every row 3 column 2/.style={highlight bold},
        every row 0 column 3/.style={highlight bold},
        every row 3 column 4/.style={highlight bold},
        every row 0 column 5/.style={highlight bold},
        every row 0 column 1/.style={postproc cell content/.style=
        {@cell content=\texttt{OS-CDA}}},
        every row 1 column 1/.style={postproc cell content/.style=
        {@cell content=\texttt{US-CDA}}},
        every row 2 column 1/.style={postproc cell content/.style=
        {@cell content=\texttt{OS-CDA}}},
        every row 3 column 1/.style={postproc cell content/.style=
        {@cell content=\texttt{US-CDA}}},
    ]
    {csv_files/jigsaw-alternative-comparison-resampling.csv}%
    \caption{Debiasing performance between two different strategies of combining resampling and CDA. The results shown are evaluated on the BERT model, averaged over 5 different runs. $\mathcal{SG}^{\mathsf{PPR}}$ and $\mathcal{CG}^{\mathsf{PPR}}$ are on a log scale with base $10^{-2}$.}
    \label{tab:jigsaw-alternative-comparison-resampling}
\end{table*}

%% file: tables/jigsaw-reweight-strategy-comparison.tex
\begin{table*}[!h]
    \centering
    \begin{filecontents*}{csv_files/jigsaw-alternative-comparison-reweight.csv}
strategy,statistical_ppr_gap_bert,statistical_ppr_gap_std_bert,causal_ppr_gap_bert,causal_ppr_gap_std_bert,statistical_ppr_gap_albert,statistical_ppr_gap_std_albert,causal_ppr_gap_albert,causal_ppr_gap_std_albert
Same weight,-2.2972949357360615,0.3505672911866903,0.16152716593245214,0.10914881605960738,-2.407564300059376,0.29707257302705126,0.06975036710719543,0.05850725936310301
Counterfactual gender weight,-1.822430491675376,0.3565078693353991,0.6534508076358301,0.24173388594940287,-2.1912747452800527,0.30825716004073583,0.37077826725403773,0.06294577165742385
Weight=1,-1.755500499154087,0.3638498699330784,0.3267254038179149,0.11037662539187147,-1.9608203602578402,0.24507246871135197,0.2386196769456679,0.09066878880490817
\end{filecontents*}
\pgfplotstabletypeset[
        col sep = comma,
        columns={strategy, sg-ppr-bert, cg-ppr-bert, sg-ppr-albert, cg-ppr-albert},
        columns/strategy/.style={column name=Strategy, string type, column type/.add={}{|}},
        columns/sg-ppr-bert/.style={string type, column name=$\mathcal{SG}^{\mathsf{PPR}}$},
        columns/cg-ppr-bert/.style={string type, column name=$\mathcal{CG}^{\mathsf{PPR}}$, column type/.add={}{|}},
        columns/sg-ppr-albert/.style={string type, column name=$\mathcal{SG}^{\mathsf{PPR}}$},
        columns/cg-ppr-albert/.style={string type, column name=$\mathcal{CG}^{\mathsf{PPR}}$},
        create on use/sg-ppr-bert/.style={
            create col/assign/.code={%
              \edef\entry{\noexpand\pgfmathprintnumber[fixed, fixed zerofill, precision=2]{\thisrow{statistical_ppr_gap_bert}}$\pm$\noexpand\pgfmathprintnumber[fixed, fixed zerofill, precision=2]{\thisrow{statistical_ppr_gap_std_bert}}}%
              \pgfkeyslet{/pgfplots/table/create col/next content}\entry
            }
          },
        create on use/cg-ppr-bert/.style={
            create col/assign/.code={%
              \edef\entry{\noexpand\pgfmathprintnumber[fixed, fixed zerofill, precision=3]{\thisrow{causal_ppr_gap_bert}}$\pm$\noexpand\pgfmathprintnumber[fixed, fixed zerofill, precision=3]{\thisrow{causal_ppr_gap_std_bert}}}%
              \pgfkeyslet{/pgfplots/table/create col/next content}\entry
            }
          },
        create on use/sg-ppr-albert/.style={
            create col/assign/.code={%
              \edef\entry{\noexpand\pgfmathprintnumber[fixed, fixed zerofill, precision=2]{\thisrow{statistical_ppr_gap_albert}}$\pm$\noexpand\pgfmathprintnumber[fixed, fixed zerofill, precision=2]{\thisrow{statistical_ppr_gap_std_albert}}}%
              \pgfkeyslet{/pgfplots/table/create col/next content}\entry
            }
          },
        create on use/cg-ppr-albert/.style={
            create col/assign/.code={%
              \edef\entry{\noexpand\pgfmathprintnumber[fixed, fixed zerofill, precision=3]{\thisrow{causal_ppr_gap_albert}}$\pm$\noexpand\pgfmathprintnumber[fixed, fixed zerofill, precision=3]{\thisrow{causal_ppr_gap_std_albert}}}%
              \pgfkeyslet{/pgfplots/table/create col/next content}\entry
            }
          },
        every head row/.style={before row={\toprule & \multicolumn{2}{c|}{BERT-Base-Uncased} & \multicolumn{2}{c}{ALBERT-Large}\\},after row=\midrule},
        every last row/.style={after row=\bottomrule},
        every row 2 column 1/.style={highlight bold},
        every row 0 column 2/.style={highlight bold},
        every row 2 column 3/.style={highlight bold},
        every row 0 column 4/.style={highlight bold},
    ]
    {csv_files/jigsaw-alternative-comparison-reweight.csv}%
    \caption{Debiasing performance of different reweighting strategies on counterfactual examples for \texttt{RW-CDA}. The results shown are evaluated on the BERT model, averaged over 5 different runs. $\mathcal{SG}^{\mathsf{PPR}}$ and $\mathcal{CG}^{\mathsf{PPR}}$ are on a log scale with base $10^{-2}$.}
    \label{tab:jigsaw-alternative-comparison-reweight}
\end{table*}